\documentclass[11pt]{article}
\usepackage{amsmath, amsthm, amsfonts, amssymb, mathrsfs, float}
\usepackage{algorithm}
\usepackage{algorithmic}
\usepackage{comment}
\usepackage{bm}
\usepackage{verbatim}
\usepackage{mathtools}
\usepackage{bbm}
\usepackage{dsfont}
\usepackage{caption}
\usepackage{graphicx}
\usepackage{graphics}
\usepackage{epstopdf}
\usepackage{tikz-cd}
\usepackage{color}

\usepackage{appendix}
\usepackage{authblk}
\usepackage{hyperref}
\hypersetup{
  colorlinks=true,
  linkcolor=blue,
  filecolor=blue,
  anchorcolor=blue,
  urlcolor=cyan,
  citecolor=purple
}

\theoremstyle{plain}
\newtheorem{theorem}{Theorem}[section]
\newtheorem{definition}[theorem]{Definition}

\newtheorem{lemma}[theorem]{Lemma}
\newtheorem{proposition}[theorem]{Proposition}
\newtheorem{corollary}[theorem]{Corollary}
\newtheorem{remark}[theorem]{Remark}

\newtheorem{exm}[theorem]{Example}

\renewcommand{\algorithmicrequire}{\textbf{Input:}}
	
\renewcommand{\algorithmicensure}{\textbf{Output:}}

\DeclareMathOperator{\argmin}{argmin}

\DeclareMathOperator{\diag}{diag}
\DeclareMathOperator{\Diag}{Diag}
\DeclareMathOperator{\dist}{dist}
\DeclareMathOperator{\disth}{\hat{d}}
\DeclareMathOperator{\Opt}{Opt}
\DeclareMathOperator{\epi}{epi}

\usepackage[textsize=tiny]{todonotes}

\title{\bf LogSpecT: Feasible Graph Learning Model from Stationary Signals with Recovery Guarantees}
\author{Shangyuan Liu\thanks{Department of Systems Engineering and Engineering Management, The Chinese University of Hong Kong. E-mail: shangyuanliu@link.cuhk.edu.hk} \qquad  Linglingzhi Zhu\thanks{Department of Systems Engineering and Engineering Management, The Chinese University of Hong Kong. E-mail: llzzhu@se.cuhk.edu.hk}\qquad 
Anthony Man-Cho So\thanks{Department of Systems Engineering and Engineering Management, The Chinese University of Hong Kong. E-mail: manchoso@se.cuhk.edu.hk}}
   
\date{\today}

\begin{document}

\maketitle

\begin{abstract}
Graph learning from signals is a core task in Graph Signal Processing (GSP). One of the most commonly used models to learn graphs from stationary signals is \textit{SpecT} \cite{segarra2017network}. However, its practical formulation \textit{rSpecT} is known to be sensitive to hyperparameter selection and, even worse, to suffer from infeasibility. In this paper, we give the first condition that guarantees the infeasibility of rSpecT and design a novel model (\textit{LogSpecT}) and its practical formulation (\textit{rLogSpecT}) to overcome this issue. Contrary to rSpecT, the novel practical model rLogSpecT is always feasible. Furthermore, we provide recovery guarantees of rLogSpecT, which are derived from modern optimization tools related to epi-convergence. These tools could be of independent interest and significant for various learning problems. To demonstrate the advantages of rLogSpecT in practice, a highly efficient algorithm based on the linearized alternating direction method of multipliers (L-ADMM) is proposed. The subproblems of L-ADMM admit closed-form solutions and the convergence is guaranteed. Extensive numerical results on both synthetic and real networks corroborate the stability and superiority of our proposed methods, underscoring their potential for various graph learning applications.
\end{abstract}

\section{Introduction}
\label{section-intro}
Learning with graphs has proved its relevance in many practical areas, such as life science \cite{smith2011network,stegle2015computational}, signal processing \cite{jung2019semi,jung2019localized,tanaka2020generalized,tanaka2020sampling}, and financial engineering \cite{acemoglu2015systemic,marti2021review}, to just name a few. However, there are many cases that the graphs are not readily prepared and only the data closely related to the graphs can be observed. Hence, a core task in Graph Signal Processing (GSP) is to learn the underlying graph topology based on the interplay between data and graphs \cite{mateos2019connecting}.

One commonly used property of data is graph signal stationarity \cite{girault2015stationary,marques2017stationary,perraudin2017stationary}. This property extends the notion of signal stationarity defined on the space/time domain to the signals with irregular structure (i.e. graphs) \cite{segarra2018statistical}. Although this concept is proposed more from a theoretical end, several works have shown that some real datasets can be approximately viewed as stationary or partially explained by the stationarity assumption. 
% For example, \cite{perraudin2017stationary} showed that the well-known USPS dataset and the CMUPIE set of cropped faces are nearly stationary. \cite{girault2015stationary} showed that a part of the weather data can be explained by stationary graph signals. \cite{segarra2017network} illustrated the relevance of stationarity assumption in learning protein structure.
For instance, \cite{perraudin2017stationary} revealed that the well-known USPS dataset and the CMUPIE set of cropped faces exhibit near stationarity. \cite{girault2015stationary} found that certain weather data could be explained by stationary graph signals. Additionally, \cite{segarra2017network} highlighted the significance of the stationarity assumption in learning protein structure which is a crucial task in bioinformatics.

The predominant methods to process stationary graph signals and learn topology under the stationarity assumption are the spectral-template-based models. The start of this line of research is \cite{segarra2017network}, which proposed a fundamental model called {\it SpecT} \cite{segarra2017network} to learn graphs from stationary graph signals. Many extensions of this fundamental model have been made since then \cite{segarra2017joint,buciulea2022learning}. In practice, SpecT requires the unknown data covariance matrix. Hence, a robust formulation called \textit{rSpecT} is proposed \cite{segarra2017network}, which introduces a hyperparameter to reflect the estimation inaccuracy of the data covariance matrix. As a consequence, the model is sensitive to this hyperparameter and improper tuning of it may jeopardize the model performance or lead to model infeasibility. 

The current approach to selecting an appropriate value is computationally costly, as it requires a highly-accurate solution to a second-order conic programming. Such an issue not only induces the unstable performance of rSpecT, but also makes the existing recovery guarantees incomplete. Another model that gains less attention tries to circumvent the model infeasibility issue by turning a constraint into a penalty \cite{shafipour2020online}. However, this approach introduces another hyperparameter that is neither easy to tune nor amenable to theoretical analysis. Also, this hyperparameter should go to infinity heuristically when more and more samples are collected, which may exacerbate model instability. Thus, we now arrive at the following natural question:
\begin{center}
\textit{Does there exist a fundamental graph learning model from stationary signals that is robust to the hyperparameters and has sound theoretical guarantees? } 
\end{center}
In this paper, we answer this essential question by proposing an alternative formulation to learn graphs without isolated nodes from stationary signals and providing rigorous recovery guarantees. More specifically, our contributions are as follows.
\subsection{Contributions}
Firstly, we provide a condition that guarantees the infeasibility of the fundamental model rSpecT. To overcome the infeasibility issue, we propose a novel formulation called \textit{LogSpecT} based on the spectral template and a log barrier to learn graphs without isolated nodes. Since LogSpecT requires the information of the unknown covariance matrix,  we introduce a practical formulation called \textit{rLogSpecT} to handle the finite-sample case. Unlike rSpecT, rLogSpecT is guaranteed to be feasible all the time. Moreover, as our proposed formulation is general enough, it can inherit almost all existing extensions of rSpecT. 

Secondly, we investigate several theoretical properties of the proposed models and connect rLogSpecT to LogSpecT via non-asymptotic recovery analysis. The recovery guarantee is proven with modern optimization tools related to epi-convergence \cite{royset2021optimization}. Different from the current guarantees for rSpecT that are built on an $\ell_1$ analysis model \cite{zhang2016one}, our approach based on epi-convergence is not limited by the type of optimization problems (e.g. the combination of the $\ell_1$-norm and log barrier in the objective function) and consequently admits broader applications that can be of independent interest.

In the algorithmic aspect, we design a linearized alternating direction method of multipliers (L-ADMM) to solve rLogSpecT. The subproblems of L-ADMM admit closed-form solutions and can be implemented efficiently due to the linearization technique. Also, we establish the convergence result of the proposed method.

Finally, extensive experiments on both synthetic data and real networks are conducted. The infeasibility issue of rSpecT is frequently observed from  experiments on real networks, demonstrating that it is more than a theoretical concern. We observe that our novel models (i.e., LogSpecT and rLogSpecT) are more accurate and stable when compared to others. For various cases where the classic model fails (e.g. BA graphs), our model can achieve excellent performance. This empirically illustrates that our model 
% is a better candidate than the classic one as a fundamental tool. 
serve as superior alternatives to classical methods for being a fundamental model.

\subsection{Related Work}
Graph learning from stationary signals is a recent popular topic.
\cite{zhu2017learning} resorted to a first-order approximation to learn graph structure from stationary signals. \cite{segarra2017network} proposed the first versatile and efficient model based on the spectral template. To deal with the issue of insufficient samples, they also introduced the robust formulation. After that, many extensions have been proposed. \cite{segarra2017joint} inferred multiple graphs jointly from stationary signals. \cite{buciulea2019network} dealt with the case where hidden nodes exist. \cite{shafipour2019online} moved to the online setting and considered streaming signals. \cite{rey2022joint} addressed the problem of joint inference with hidden nodes. \cite{shafipour2020online} considered the online setting with hidden nodes. \cite{buciulea2022learning} learned graphs with hidden variables from both stationary and smooth signals. 

Different from these extensions of settings, we revisit the basic model and pose regularizers on the degrees. As discussed in \cite{de2021graphical}, restricting the degrees to preclude isolated nodes is crucial to learning graphs with special structures. However, the direct adoption of constraints may even aggravate the infeasibility issue in our problem. We resort to the log barrier as a degree regularization term.
In graph learning, the log barrier first appeared in \cite{kalofolias2016learn} to learn graphs from smooth signals. The follow-up works are confined to learning from smooth signals in different settings \cite{kalofolias2017learning,maretic2020graph,zheng2022multi}. To the best of our knowledge, this is the first work that demonstrates, both theoretically and empirically, the relevance of the log barrier in graph learning from stationary signals.

Though the robust formulations are ubiquitous in various models of graph learning from stationary signals, their recovery guarantees are rarely studied. \cite{pasdeloup2017characterization}  empirically showed that the feasible set of the robust formulation approximates the original one as more samples are collected.  \cite{navarro2020joint} provided the first recovery guarantee for rSpecT. Their work relies heavily on the fact that the objective is the $\ell_1$-norm and the constraints are linear \cite{zhang2016one}. In particular, their approach is not applicable to our model. Moreover, the conditions needed in their work 
are not only restrictive but also hard to check as they depend on the full rankness of a large-scale matrix related to random samples.

\subsection{Notation}
The notation we use in this paper is standard. We use $[m]$ to denote the set $\{1,2,\ldots,m\}$ for any positive integer $m$. Let the Euclidean space of all real matrices be equipped with the inner product $\langle \bm{X}, \bm{Y}\rangle\coloneqq\operatorname{tr}(\bm{X}^{\top} \bm{Y})$ for any matrices $\bm{X}, \bm{Y}$ and denote the induced Frobenius norm by $\|\cdot\|_F$ (or $\Vert\cdot\Vert_2$ when the argument is a vector). For any $\bm{X}\in \mathbb{R}^{m\times n}$, we use $\mathbb{B}(\bm{X},\rho)$ to denote the closed Euclidean ball centering at $\bm{X}$ with radius $\rho$. Let  $\|\cdot\|$ be the operator norm,  $\|\bm{X}\|_{1,1}\coloneqq\sum_{i,j}|\bm{X}_{ij}|$, $\|\bm{X}\|_{\infty,\infty}\coloneqq\max_{i,j}|\bm{X}_{ij}|$, and let $\diag(\bm{X})$ be vector formed by the diagonal entries of $\bm{X}$. 
For a column vector $\bm{x}$, let $\Diag(\bm{x})$ be the diagonal matrix whose diagonal elements are given by $\bm{x}$. Given a closed and convex set $\mathcal{C}$, we use $\Pi_{\mathcal{C}}(\bm{w}) \coloneqq \argmin_{\bm{v}\in\mathcal{C}}\|\bm{v}-\bm{w}\|_2$ to denote the projection of the point $\bm{w}$ onto the set $\mathcal{C}$. We use $\bm{1}$ (resp.\ $\bm{0}$) to denote an all-one vector (resp. all-zero vector) whose dimension will be clear from the context.  For a set $\mathcal{S}$ and any real number $\alpha$, let $\alpha \mathcal{S} \coloneqq \{\alpha x \mid x \in \mathcal{S}\}$, $\mathcal{S}\bm{1} \coloneqq \{\bm{x}\bm{1} \mid \bm{x} \in \mathcal{S}\}$, and $\iota_\mathcal{S}(\cdot)
% = \left\{\begin{aligned}
%     &0,  &&\text{if } \bm{x} \in \mathcal{S} \\
%     &+\infty. &&\text{if } \bm{x} \notin \mathcal{S}
% \end{aligned}\right.
$ be the indicator function of $\mathcal{S}$. For two non-empty and compact sets $\mathcal{X}$ and $\mathcal{Y}$, the distance between them is defined as $\text{dist}(\mathcal{X},\mathcal{Y}) \coloneqq \sup_{\bm{X} \in \mathcal{X}} \inf_{\bm{Y} \in \mathcal{Y}} \|\bm{X} - \bm{Y}\|_F$\footnote{It reduces to the classic point-to-set (resp. point-to-point) distance when the set $\mathcal{Y}$ (resp. $\mathcal{X}$ and $\mathcal{Y}$) is a singleton.
}.
% is used to denote the distance between $\mathcal{A}$ and $\mathcal{B}$. 

% {\color{red} dist}\\

\section{Preliminaries}
\label{sec-pre}
Suppose that $\mathcal{G} = (\mathcal{V}, \mathcal{E})$ is a graph, where $\mathcal{V} = [m]$ is the set of nodes and $\mathcal{E} \subset \mathcal{V} \times \mathcal{V}$ is the set of edges. 
Let $\bm{S}$ be the weight matrix associated with the graph $\mathcal{G}$, where ${\bm S}_{ij}$ represents the weight of the edge between nodes $i$ and $j$.
In this paper, we consider undirected graphs without self-loops. Then, the set of valid adjacency matrices is \[\mathcal{S} \coloneqq \{\bm{S} \in \mathbb{R}^{m\times m} \mid \bm{S} = \bm{S}^\top,\  \diag(\bm{S}) = \bm{0},\ \bm{S} \geq 0\}.\]

Suppose that the adjacency matrix $\bm{S}$ admits the eigen-decomposition $\bm{S} = \bm{U}\bm{\Lambda}\bm{U}^\top$, where $\bm{\Lambda}$ is a diagonal matrix and $\bm{U}$ is an orthogonal matrix. A graph filter is a linear operator $h: \mathbb{R}^{m\times m}\rightarrow \mathbb{R}^{m\times m}$ defined as 
\[h(\bm{S}) = \sum_{i = 0}^{p}h_i\bm{S}^i = \bm{U}\left(\sum_{i = 0}^{p}h_i\bm{\Lambda}^i\right)\bm{U}^\top,\]
where $p>0$ is the order of the graph filter and $\{h_i\}_{i=0}^p$ are the filter coefficients. 
According to the convention, we have $\bm{S}^0 = \bm{I}$. 
% \Lingzhi{why?}. 

A graph signal can be represented by a vector $\bm{x}\in \mathbb{R}^m$, where the $i$-th element $x_i$ is the signal value associated with node $i$. A signal $\bm{x}$ is said to be stationary if it is generated from
\begin{equation}\label{equa-generative-filter}
    \bm{x} = h(\bm{S})\bm{w},
\end{equation}
where $\bm{w}$ satisfies $\mathbb{E}[\bm{w}] = \bm{0}$ and $\mathbb{E}[\bm{w}^\top\bm{w}] = \bm{I}$. Simple calculations give that the covariance matrix of $\bm{x}$, which is denoted by $\bm{C}_{\infty}$, shares the same eigenvectors with $\bm{S}$. Hence, we have the constraint
\begin{equation}\label{equa-spect}
    \bm{C}_{\infty}\bm{S} = \bm{S}\bm{C}_{\infty}.
\end{equation} Based on this, the following fundamental model SpecT is proposed to learn graphs from stationary signals without the knowledge of graph filters \cite{segarra2017network, shafipour2020online}:
\begin{equation}\label{model-class-ideal}\tag{SpecT}
	\begin{split}
		&\min_{\bm{S}}\ \|\bm{S}\|_{1,1} \\
		&\ \text{s.t.}\ \  \, \bm{C}_\infty\bm{S} = \bm{S}\bm{C}_\infty, \\
		&\ \ \ \ \ \ \ \, \bm{S} \in \mathcal{S}\cap \left\{\bm{S}\in \mathbb{R}^{m\times m} : \sum_{j=1}^{m}\bm{S}_{1j} = 1\right\},
	\end{split}
\end{equation}
where the constraint $\sum_{j=1}^{m}\bm{S}_{1j} = 1$ is used to preclude the trivial optimal solution $\bm{S}^* = \bm{0}$. When $\bm{C}_\infty$ is unknown and only $n$ i.i.d samples of $\{\bm{x}_i\}$ from \eqref{equa-generative-filter} are available, the robust formulation, which is based on the estimate $\bm{C}_n$ of $\bm{C}_\infty$ and called \textit{rSpecT}, is used:
\begin{equation}\label{model-class-noisy}\tag{rSpecT}
	\begin{split}
		&\min_{\bm{S}}\ \|\bm{S}\|_{1,1} \\
		&\ \text{s.t.}\  \, \, \|\bm{C}_n\bm{S} - \bm{S}\bm{C}_n\|_F \leq \delta, \\
		&\ \ \ \ \ \  \ \, \bm{S} \in \mathcal{S}\cap \left\{\bm{S}\in \mathbb{R}^{m\times m} : \sum_{j=1}^{m}\bm{S}_{1j} = 1\right\}.
	\end{split}
\end{equation}
For this robust formulation, the recovery guarantee is studied empirically in \cite{pasdeloup2017characterization} and theoretically in \cite{navarro2020joint} under some conditions that are restrictive and hard to check.
% \Lingzhi{Is there any proved theoretical guarantee for (3) as a robust formulation of (2) by others?}
% In the two models, the constraint $\sum_{j=1}^{m}{\bm S}_{1j} = 1$ is to preclude trivial optimal solution $\bm{S}^* = \bm{0}$.
\section{Infeasibility of rSpecT and Novel Models}
\label{section-infea}
Even though rSpecT has gained much popularity, there are few works that discuss the choice of the hyperparameter $\delta$ and the infeasibility issue. In this section, we present a condition under which rSpecT is guaranteed to be infeasible and then propose an alternative formulation. To motivate our results, let us consider the following 2-node example. 
\begin{exm}\label{exm-infeas-2nodes}
Consider a simple graph containing 2 nodes. Then, the set given by the second constraint of rSpecT is  \[\mathcal{S}\cap \left\{\bm{S}\in \mathbb{R}^{m\times m}: \sum_{j=1}^{m}\bm{S}_{1j} = 1\right\}=\left[0, 1; 1, 0\right],\] 
which is a singleton. 
% and the unique solution is: $\bm{S} = \left[0, 1; 1, 0\right]$. 
% After vectorization, we get $\text{vec}(\bm{S}) = [0,1,1,0]^\top$. 
Suppose that the sample covariance matrix is $\bm{C}_n = \left[
	h_{11}, h_{12};
	h_{12}, h_{22}\right]$. 
%  Simple calculation shows that:
% \begin{align*}
%  \bm{A} = \left[\begin{matrix}
% 		0 & h_{12} & -h_{12} & 0 \\
% 		h_{12} & h_{22}-h_{11} & 0 &-h_{12} \\
% 		-h_{12} & 0 & h_{11} - h_{22} & h_{12} \\
% 		0 & -h_{12} & h_{12} & 0
% 	\end{matrix}\right],
% \end{align*}
% where 
Then, the constraint $\|\bm{C}_n\bm{S} - \bm{S}\bm{C}_n\|_F \leq \delta$ is reduced to
% It implies that 
$2(h_{11}-h_{22})^2 \leq \delta^2$. Hence, when $h_{11} \neq h_{22}$ and $\delta<\sqrt{2}\left|h_{11}-h_{22}\right|$, rSpecT has no feasible solution.
\end{exm}
Before delving into the general case, we introduce a linear operator $\bm{B} \in \mathbb{R}^{m^2\times m(m-1)/2}$ that maps the vector $\bm{x} \in \mathbb{R}_+^{(m-1)m/2}$ to the vectorization of an adjacency matrix $\bm{S} \in \mathcal{S}$ of a simple, undirected graph:
\begin{equation}
    (\bm{B}\bm{x})_{m(i-1)+j} = 
    \left\{
    \begin{aligned}
     &x_{i-j+\frac{j-1}{2}(2m-j)}, &&\text{if} \ \ i>j, \\
     &0, &&\text{if} \ \ i = j,\\
     &x_{j-i+\frac{i-1}{2}(2m-i)},  &&\text{if} \ \ i<j,
    \end{aligned}
    \right.
\end{equation}
where 
% $\bm{B} \in \mathbb{R}^{m^2\times m(m-1)/2}$, 
$i,j \in [m]$. We also define $\bm{A}_n \coloneqq \bm{I} \otimes \bm{C}_n - \bm{C}_n \otimes \bm{I} \in \mathbb{R}^{m^2 \times m^2}$, so that the first constraint of rSpecT can be rewritten as
\begin{align*}
	&\|\bm{C}_n\bm{S} - \bm{S}\bm{C}_n\|_F \leq \delta \
	% \Leftrightarrow &\|(\bm{I} \otimes \bm{C}_n - \bm{C}_n \otimes \bm{I})\text{vec}(\bm{S})\|_F \leq \delta \\
	\Longleftrightarrow\   
	\left\|\bm{A}_n \text{vec}(\bm{S})\right\|_2 \leq \delta, 
	% \Leftrightarrow 2(h_{11}-h_{22})^2 \leq \delta^2.
\end{align*}
where 
% $\bm{A} \coloneqq \bm{I} \otimes \bm{C}_n - \bm{C}_n \otimes \bm{I}$ and 
vec$(\cdot)$ is the vectorization operator. We now give a condition that guarantees the infeasibility of rSpecT. 
% The result is formally stated as follows.

\begin{theorem}\label{thm-infeasibility}
    Consider the linear system
    \begin{equation}\label{thm-infea-linear}
        \begin{split}
            \bm{A}_n\bm{B}\bm{y} = \bm{0},\ \
            \bm{y} \leq \bm{0},\ \
         \sum_{i=1}^{m-1}y_i\neq \bm{0},
        \end{split}
    \end{equation}
    where $\bm{y}\in \mathbb{R}^{\frac{m(m-1)}{2}}$. If \eqref{thm-infea-linear} has no feasible solution, then there exists a $\bar{\delta}_n > 0$ such that rSpecT is infeasible for all $\delta_n\in [0,\bar{\delta}_n)$.
\end{theorem}
\begin{remark}\label{remark-infea}
    From Theorem \ref{thm-infeasibility} we can infer that for any fixed $n$, the linear system \eqref{thm-infea-linear} has no solution when $\bm{A}_n \bm{B}$ has full column rank (e.g., Example \ref{exm-infeas-2nodes}). This leads to the infeasibility of rSpecT with $\delta_n\in [0,\bar{\delta}_n)$. 
    % However, the probability that rSpecT will be infeasible is still unclear. 
    As remarked in \cite{segarra2017network}, one should tackle the feasibility issue of rSpecT with caution.
    % It is only empirically known that this event should be tackled with caution in practice . 
    % We leave this for the future direction.
\end{remark}
The failure of rSpecT (SpecT) lies in the existence of the constraint $(\bm{S}\bm{1})_1 = 1$, which is used to preclude the trivial solution $\bm{S}^* = \bm{0}$. For this reason, we resort to an alternative approach to bypassing the trivial solution. When the graphs are assumed to have no isolated node, the log barrier is commonly applied \cite{kalofolias2016learn}. In these cases, the zero solution is naturally precluded. This observation inspires us to propose the following novel formulation, which combines the log barrier with the spectral template \eqref{equa-spect} to learn graphs without isolated nodes from stationary signals:
\begin{equation}\label{model-novel-ideal}\tag{LogSpecT}
	\begin{split}
		&\min_{\bm{S}}\  \|\bm{S}\|_{1,1} - \alpha\bm{1}^\top\log (\bm{S1}) \\
		&\ \text{s.t.}\ \  \bm{S} \in \mathcal{S},\ \ \bm{S}\bm{C}_\infty = \bm{C}_\infty\bm{S},
	\end{split}
\end{equation}
% which is formally called \textit{LogSpecT}. 
where $\|\cdot\|_{1,1}$ is a convex relaxation of $\|\cdot\|_0$ promoting graph sparsity, $\bm{1}^\top\log (\bm{S1})$ is the penalty to guarantee nonexistence of isolated nodes, and $\alpha$ is the tuning parameter.
As can be seen from the following proposition, the hyperparameter $\alpha$ in LogSpecT
only affects the scale of edge weights instead of the graph connectivity structure.

\begin{proposition}\label{theorem-choice-parameters}
    Let $\Opt(\bm{C}, \alpha)$ be the optimal solution set of LogSpecT with input covariance matrix $\bm{C}$ and parameter $\alpha > 0$. Then, for any $\gamma > 0$, it follows that
    \begin{equation*}
        \Opt(\bm{C},\alpha) = \gamma \Opt(\bm{C}, \alpha/\gamma) = \alpha \Opt(\bm{C}, 1).
    \end{equation*}
\end{proposition}
\begin{remark}
    The result of Proposition \ref{theorem-choice-parameters} spares us from tuning the hyperparameter $\alpha$ when we are coping with binary graphs. In fact, certain normalization will eliminate the impact of different values of $\alpha$ and preserve the connectivity information. Hence, we may simply set $\alpha = 1$ in implementation.
\end{remark}
Note that the true covariance matrix $\bm{C}_\infty$ in LogSpecT is usually unknown and an estimate $\bm{C}_n$ from $n$ i.i.d samples is available. To tackle the estimate inaccuracy, we introduce the following robust formulation:
% In this case, the robust formulation is introduced to tackle the estimation inaccuracy as follows:
\begin{equation}\label{model-novel-con}\tag{rLogSpecT}
	\begin{split}
		&\min_{\bm{S}}\  \|\bm{S}\|_{1,1} - \alpha\bm{1}^\top\log (\bm{S1}) \\
		&\ \text{s.t.}\ \ \bm{S} \in \mathcal{S}, \ \ \|\bm{S}\bm{C}_n - \bm{C}_n\bm{S}\|_{F}^2 \leq \delta_n^2.
	\end{split}
\end{equation}
This formulation substitutes $\bm{C}_\infty$ with $\bm{C}_n$ and relaxes the equality constraint to an inequality constraint with a tolerance threshold $\delta_n$. Contrary to rSpecT, we prove that rLogSpecT is always feasible.
\begin{proposition}\label{prop-feasible}
    For  
    % $\alpha > 0$, 
    $\delta_n > 0$ and $\bm{C}_n$ with any fixed $n$, rLogSpecT always has a nontrivial feasible solution. 
\end{proposition}

Next, we discuss some properties of the optimal solutions/value of the proposed models, which are useful for deriving the recovery guarantee. More specifically, we obtain an upper bound on the optimal solutions (which may not be unique) independent of the sample size $n$ and the inaccuracy parameter $\delta_n$. Also, a lower bound of optimal values follows.

% Recall that $m$ is the number of nodes.
\begin{proposition}\label{lemma-bound-solutions}
    The following statements hold:
    \begin{itemize}
    \item For an optimal solution $\bm{S}^*$ (resp. $\bm{S}_n^*$) to LogSpecT (resp. rLogSpecT with any given sample size $n$), it follows that
    \begin{align*}
        \|\bm{S}^*\|_{1,1} = \alpha m \ \ \text{and}\ \
        \|\bm{S}_n^*\|_{1,1} \leq \alpha m, \ \ \forall \delta_n > 0.
    \end{align*}
    % where  
    \item If $\delta_n \geq 2\alpha m\|\bm{C}_n-\bm{C}_\infty\|$, then
\begin{equation*}
    \alpha m(1-\log \alpha) \leq f^*_n \leq f^*, \ \ \forall n,
\end{equation*}
where $f^*$ (resp. $f^*_n$) denotes the optimal value of LogSpecT (resp. rLogSpecT). 
    \end{itemize}
    % $\alpha$ is the model parameter, $m$ is the number of nodes.
\end{proposition}

% The second result is about the bounds of the optimal values.
% \begin{lemma}\label{lemma-bound-values}
% If $\delta_n \geq 2\alpha m\|\bm{C}_n-\bm{C}_\infty\|$, then
% \begin{equation*}
%     \alpha m(1-\log \alpha) \leq v^*_n \leq v^*, \ \ \forall n.
% \end{equation*}
% where $v^*$ (resp. $v^*_n$) is the optimal value of LogSpecT (resp. rLogSpecT). 
% \end{lemma}
% Note that $v^*$ is a finite value that is irrelevant of $n$. Hence, the above lemma provides sample-size-free bounds of $v_n^*$.

% Similarly to \eqref{model-class-noisy}, the recovery performance of model \eqref{model-novel-con} can be guaranteed theoretically. This part is discussed in the next section.

\section{Recovery Guarantee of rLogSpecT}
\label{section-recovery}
In this section, we investigate the non-asymptotic behavior of rLogSpecT when more and more i.i.d samples are collected. For the sake of brevity, we denote 
$f \coloneqq \|\cdot\|_{1,1} - \alpha \bm{1}^\top \log \left(\cdot\bm{1}\right)$. 
The theoretical recovery guarantee is as follows:
\begin{theorem}\label{thm-main}
    If $\delta_n \geq 2\rho\|\bm{C}_n - \bm{C}_\infty\|$ with given $\rho \coloneqq \max\{f^*, \alpha m, \alpha m(\log \alpha -1)\}$, then there exist constants $c_1,\Tilde{c}_1, \Tilde{c}_2 > 0$ such that
    \begin{enumerate}
        \item[{\rm (i)}] $|f_n^* - f^*| \leq c_1 \varepsilon_n$, \label{theorem2-f}
        \item[{\rm (ii)}] ${\rm dist}(\mathcal{S}^{n,*}, \mathcal{S}^*_{2\varepsilon_n}) \leq \varepsilon_n$, \label{theorem2-subopt}
        \item[{\rm (iii)}] ${\rm dist}(\mathcal{S}^{n,*}\bm{1}, \mathcal{S}^*_{0}\bm{1}) \leq \Tilde{c}_1\varepsilon_n + \Tilde{c}_2\sqrt{\varepsilon_n}$,\label{theorem2-opt}
    \end{enumerate}
    % \[|f_n - f^*| \leq c\cdot \varepsilon_n\quad \text{and}\quad {\rm dist}(\mathcal{S}^{n,*}, \mathcal{S}^*_{2\varepsilon_n}) \leq \varepsilon_n,\] 
    where $\varepsilon_n \coloneqq \delta_n + 2\rho\|\bm{C}_n - \bm{C}_\infty\|$,
    $\mathcal{S}^{n,*}$ is the optimal solution set of rLogSpecT, and 
    \[\mathcal{S}^*_{\varepsilon}\coloneqq\{\bm{S}\in \mathcal{S} \mid \bm{S}\bm{C}_\infty = \bm{C}_\infty\bm{S},\ f(\bm{S}) \leq f^*+\varepsilon\}\] is the $\varepsilon$-suboptimal solution set of LogSpecT.
\end{theorem}

\begin{remark}
   {\rm (i)} Compared with the conclusion {\rm (ii)} in the Theorem \ref{thm-main}, the conclusion {\rm (iii)} links the optimal solution sets of rLogSpecT and LogSpecT instead of the sub-optimal solutions.\\ 
   {\rm (ii)}  A byproduct of the proof (see Appendix \ref{appen-proof-thm-main} for details) shows that the optimal node degree vector set ($\mathcal{S}_0^*\bm{1}$) is a singleton. However, there is no guarantee of the uniqueness of $\bm{S}^*$ itself. We will discuss the impact of such non-uniqueness on model performance. This will be discussed in Section \ref{exp-syn}.
    % {\rm (iii)} The recovery guarantees for rSpecT in \cite{navarro2020joint} require the full rankness of a matrix. This is restrictive and hard to check since the relevant matrix is extremely large and depends on samples. In comparison, our recovery guarantees barely need assumptions on samples.
\end{remark}

\begin{remark}
% Another line of work related to this paper is the optimization tools we are resorting to. 
The proof of Theorem \ref{thm-main} relies on two important optimization concepts: Epi-convergence and truncated Hausdorff distance.
Epi-convergence is closely related to the asymptotic solution behavior of approximate minimization problems, 
% It characterizes the convergence property of approximations to the actual model which guarantees approximate solutions converge to an actual solution. 
% The appearance of epi-convergence goes back to the pioneering work by \cite{wijsman1964convergence} and \cite{mosco1969convergence}, and the term epi-convergence was first used by \cite{wets1980convergence}. 
% % with an increasing focus on nonconvex optimization. 
% Epi-convergence coincides with the $\Gamma$-convergence in most cases, which was independently developed by Di Giorgi and Franzoni \cite{de1975tipo} in their work on the calculus of variation. 
and the truncated Hausdorff distance is used to characterize the epi-convergence non-asymptotically. With the help of the Kenmochi condition,  which allows us to explicitly calculate the truncated Hausdorff distance, 
% characterized by the Kenmochi condition, 
we are able to study the non-asymptotic behaviour of the optimal value and optimal solutions of the models. We refer the reader to Appendix \ref{appendix-main} and also
\cite[Chapter 7]{rockafellar2009variational},
\cite[Chapter 6.J]{royset2021optimization} for more details. 
\end{remark}

\begin{corollary}\label{coro-conv-sol}
    Under the assumptions in Theorem \ref{thm-main}, it follows that if  $\delta_n\rightarrow0$, then
    % \begin{itemize}
        % \item if  $\delta_n\rightarrow0$ then
        % and $\|\bm{C}_n - \bm{C}_{\infty}\| \rightarrow 0$,
        \begin{equation}\label{coro-conv-sol-1}
            \lim_{n\rightarrow \infty}\dist(\mathcal{S}^{n,*}, \mathcal{S}^*_0) = 0.
            % \quad \text{if}\ \ \delta_n\rightarrow0.
        \end{equation}
\end{corollary}
\begin{remark}
  From the strong law of large numbers, if we choose $\delta_n = \mathcal{O}(\|\bm{C}_n-\bm{C}_\infty\|)$, then  
    % when $n\rightarrow\infty$ and $\delta_n = \mathcal{O}(\|\bm{C}_\infty - \bm{C}_n\|)$, 
    almost surely one has 
    % that 
% \begin{equation*}
$\lim_{n\rightarrow \infty}\|\bm{C}_n-\bm{C}_\infty\|=0$
% \end{equation*}
and consequently $\delta_n\rightarrow 0$.
Hence, Corollary \ref{coro-conv-sol} shows that \eqref{coro-conv-sol-1} holds almost surely.
\end{remark}

% In this paper, we c

For the remaining part of this section, we investigate the choice of $\delta_n$ under certain statistical assumptions. A large number of distributions (e.g., Gaussian distributions, exponential distributions and any bounded distributions) can be covered by the sub-Gaussian distribution (cf. \cite[Section 2.5]{vershynin2018high}), whose formal definition is as follows.
\begin{definition}[Sub-Gaussian Distributions]
     The probability distribution of a random vector $\bm{w}$ is called sub-Gaussian if there are positive constants $C, v$ such that for every $t > 0$,
     \begin{equation*}
         \mathbb{P}(\|\bm{w}\|_2>t)\leq Ce^{-vt^2}.
     \end{equation*}
\end{definition}

 Consider the case that $\bm{w}$ in the generative model \eqref{equa-generative-filter} follows a sub-Gaussian distribution. The following result is adapted from \cite[Proposition 2.1]{vershynin2012close}.
\begin{lemma}\label{lemma-subgaussian}
Suppose that $\bm{w}$ in the generative model \eqref{equa-generative-filter} follows a sub-Gaussian distribution. Then, $\bm{x}$ follows a sub-Gaussian distribution, and with probability larger than $1-\mathcal{O}(\frac{1}{n})$,
\begin{equation*}
	\left\|\bm{C}_n - \bm{C}_\infty\right\| \leq \mathcal{O}\left(\sqrt{\frac{\log n}{n}}\right).
\end{equation*} 
\end{lemma}

% Furthermore, Theorem \ref{thm-main} indicates that $\delta_n$ should decay in the same order with $\|\bm{C}_n-\bm{C}_\infty\|$. Hence, we may choose $\delta_n = \mathcal{O}(\left(\log n/n\right)^{\frac{1}{2}})$.

Equipped with the non-asymptotic results in Theorem \ref{thm-main}, we can choose $\delta_n$ with a specific decaying rate.

\begin{corollary}\label{main-cor}
If the input $\bm{w}$ in \eqref{equa-generative-filter} follows a sub-Gaussian distribution and $\delta_n = \mathcal{O}(\sqrt{\log n/n})$, then the assumptions in Theorem \ref{thm-main} hold with probability larger than $1-\mathcal{O}(\frac{1}{n})$.
\end{corollary}

Corollary \ref{main-cor} together with Theorem \ref{thm-main} illustrates the non-asymtotic convergence of optimal function value/suboptimal solution set/optimal node degree vector of rLogSpect from $n$ i.i.d samples to the ideal model LogSpect with the convergence rate $\mathcal{O}(\varepsilon_n) / \mathcal{O}(\varepsilon_n) / \mathcal{O}(\sqrt{\varepsilon_n})$, where $\varepsilon_n=\mathcal{O}\left(\left|\bm{C}_n-\bm{C}_{\infty}\right|\right) \leq \mathcal{O}(\sqrt{\log n / n})$ for sub-Gaussian $\omega$. This matches the convergence rate $\mathcal{O}(\sqrt{\log n / n})$ of classic spectral template models, e.g., Proposition 2 in \cite{segarra2017network} and Theorem 2 in \cite{navarro2020joint} for the SpecT model, which shows that LogSpecT is also competitive with respect to the recovery guarantee.

\section{Linearized ADMM for Solving rLogSpecT}
\label{section-algo}
 In this section, we design a linearized ADMM algorithm to solve rLogSpecT that admits closed-form solutions for the subproblems. The ADMM-type algorithms have been successfully applied to tackle various graph learning tasks \cite{zhao2019admmforgraph,wang2021efficient}. 
 
 Firstly, we reformulate rLogSpecT such that it fits the ADMM scheme:
\begin{equation}\label{model-reformulation}
	\begin{split}
		&\min_{\bm{S},\bm{Z},\bm{q}}\  \langle \bm{S}, \bm{1}\bm{1}^\top \rangle - \alpha \bm{1}^\top\log \bm{q} + \iota_{\mathcal{S}}(\bm{S}) + \iota_{\mathbb{B}(\bm{0},\delta_n)}(\bm{Z}) \\
		&\ \ \text{s.t.}\  \ \bm{C}_n\bm{S}-\bm{S}\bm{C}_n= \bm{Z}, \ \ \bm{q} = \bm{S1}.
	\end{split}
\end{equation}
The augmented Lagrangian function of problem \eqref{model-reformulation} is
\begin{equation*}
	\begin{split}
      &L(\bm{S},\bm{Z},\bm{q},\bm{\Lambda},\bm{\lambda}_2)\\
      =\ &\langle \bm{11}^\top,\bm{S} \rangle -\alpha \bm{1}^\top\log \bm{q} + \iota_{\mathcal{S}}(\bm{S}) + \iota_{\mathbb{B}(\bm{0},\delta_n)}(\bm{Z})+\langle \bm{\Lambda}, \bm{C}_n\bm{S} - \bm{S}\bm{C}_n-\bm{Z} \rangle + \bm{\lambda}_2^\top \left(\bm{q}-\bm{S1}\right) \\
		&+ \frac{\rho}{2}\|\bm{C}_n\bm{S}-\bm{S}\bm{C}_n-\bm{Z}\|_F^2 + \frac{\rho}{2}\|\bm{q}-\bm{S1}\|_2^2.
	\end{split}	
\end{equation*}
Since $\bm{q}$ and $\bm{Z}$ are separable given $\bm{S}$, we update the primal variables in two blocks: $\bm{S}$ and $\left(\bm{Z},\bm{q} \right)$. More specifically, the update of $\left(\bm{Z},\bm{q} \right)$ in the $k$-th iteration is
\begin{equation}\label{algo-update-zq}
    ( \bm{Z}^{(k+1)}, \bm{q}^{(k+1)}) \coloneqq \underset{(\bm{Z},\bm{q})}{\argmin}\, L(\bm{S}^{(k)},\bm{Z},\bm{q},\bm{\Lambda}^{(k)},\bm{\lambda}_2^{(k)}),
\end{equation}
which admits the following closed-form solution:
\begin{proposition}\label{prop-algo-sub}
    The update \eqref{algo-update-zq} can be explicitly rewritten as
    % admits a closed-form solution:
    \begin{align*}
    % &\tilde{\bm{z}} = \bm{C}_n\bm{S}^{(k)}-\bm{S}^{(k)}\bm{C}_n+\frac{\bm{\Lambda}^{(k)}}{\rho}, \bm{Z}^{(k+1)} = \min\{1, \frac{\delta_n}{\|\Tilde{\bm{Z}}\|_F}\}\tilde{\bm{z}}, \\
    % &\bm{v} = \frac{1}{\rho}\left(\rho\bm{S}^{(k)}\bm{1}-\bm{\lambda}_2^{(k)}\right),
    % \bm{q}^{(k+1)} = \frac{\bm{v}+\sqrt{\bm{v}^2+4\alpha/\rho\bm{1}}}{2}.\\
    \bm{Z}^{(k+1)} &= \min\left\{1, \frac{\delta_n}{\|\Tilde{\bm{Z}}\|_F}\right\}\tilde{\bm{Z}}, \\
    \bm{q}^{(k+1)} &= \frac{\tilde{\bm{q}}+\sqrt{\tilde{\bm{q}}^2+4\alpha/\rho\bm{1}}}{2},
    \end{align*}
where
    $\tilde{\bm{Z}} = \bm{C}_n\bm{S}^{(k)}-\bm{S}^{(k)}\bm{C}_n+\frac{\bm{\Lambda}^{(k)}}{\rho}$, $\tilde{\bm{q}} = \bm{S}^{(k)}\bm{1}-\frac{1}{\rho}\bm{\lambda}_2^{(k)}$.
\end{proposition}
The ADMM update of $\bm{S}$ does not admit a closed-form solution. Hence, we solve the linearized version $\hat{L}_k$ of the augmented Lagrangian function $L$ when all the variables except $\bm{S}$ are fixed, which is given by
\begin{equation*}
\hat{L}_k(\bm{S}) = \langle \bm{C}^{(k)},\bm{S}\rangle  +\frac{\rho\tau}{2}\|\bm{S}-\bm{S}^{(k)}\|_F^2 + \iota_{\mathcal{S}}(\bm{S}),
\end{equation*}
where $\bm{C}^{(k)} 
\coloneqq \bm{11}^\top+\bm{C}_n\bm{\Lambda}^{(k)}-\bm{\Lambda}^{(k)}\bm{C}_n-\bm{\lambda}_2^{(k)}\bm{1}^\top+\rho (\bm{S}^{(k)}\bm{C}_n^2+\bm{C}_n^2\bm{S}^{(k)}-2\bm{C}_n\bm{S}^{(k)}\bm{C}_n+\bm{Z}^{(k+1)}\bm{C}_n-\bm{C}_n\bm{Z}^{(k+1)}-\bm{q}^{(k+1)}\bm{1}^\top+ \bm{S}^{(k)}\bm{1}\bm{1}^\top )$.
% \begin{align*}
% \bm{C}^{(k)} 
% \coloneqq\ & \bm{11}^\top+\bm{C}_n\bm{\Lambda}^{(k)}-\bm{\Lambda}^{(k)}\bm{C}_n-\bm{\lambda}_2^{(k)}\bm{1}^\top\\
% &+\rho (\bm{S}^{(k)}\bm{C}_n^2+\bm{C}_n^2\bm{S}^{(k)}-2\bm{C}_n\bm{S}^{(k)}\bm{C}_n\\
% &+\bm{Z}^{(k+1)}\bm{C}_n-\bm{C}_n\bm{Z}^{(k+1)}-\bm{q}^{(k+1)}\bm{1}^\top+ \bm{S}^{(k)}\bm{1}\bm{1}^\top )
% \end{align*}
Therefore, the update of $\bm{S}$ in the $k$-th iteration is
\begin{align}\label{algo-update-S}
    \bm{S}^{(k+1)} &\coloneqq \underset{\bm{S}\in \mathcal{S}}{\argmin}\, \hat{L}_k(\bm{S}) = \Pi_{\mathcal{S}}(\bm{S}^{(k)} - \bm{C}^{(k)}/\rho\tau ),
\end{align}
where $\Pi_{\mathcal{S}}(\bm{X})$ is the projection of $\bm{X}$ onto the closed and convex set $\mathcal{S}$ and can be computed by
% efficiently with Algorithm \ref{algo-p2s}.
\begin{equation*}
    (\Pi_{\mathcal{S}}(\bm{X}))_{ij} = 
    \left\{
    \begin{aligned}
     &\frac{1}{2}\max\{0, X_{ij}+X_{ji}\}, &&\text{if} \ \ i\neq j, \\
     &0, &&\text{if} \ \ i = j.\\
     % &x_{j-i+\frac{i-1}{2}(2m-i)},  &&\text{if} \ \ i<j
    \end{aligned}
    \right.
\end{equation*}

% \begin{algorithm}[H]
% 	\caption{Projection onto $\mathcal{S}$}
% 	\begin{algorithmic}[1]\label{algo-p2s}
% 		\REQUIRE $\bm{S}$
% 		\ENSURE $\Pi_{\mathcal{S}}(\bm{S})$ 
% 		\STATE {$\bm{S} = \frac{1}{2}\left(\bm{S}+\bm{S}^\top\right)$;}
% 		\STATE{$\diag(\bm{S}) = 0$;}
% 		\STATE{$S_{ij} = \max\{0,S_{ij}\}, \forall i,j = 1,\ldots,m$;}
% 		\STATE{$\Pi_{\mathcal{S}}(\bm{S}) = \bm{S}$.}
% 	\end{algorithmic} 
% \end{algorithm}

Finally, we update the dual variables $\bm{\Lambda}$ and $\bm{\lambda}_2$ via gradient ascent steps of the augmented Lagrangian function:
\begin{align}
    \bm{\Lambda}^{(k+1)} &\coloneqq \bm{\Lambda}^{(k)} + \rho(\bm{C}_n\bm{S}^{(k+1)}-\bm{S}^{(k+1)}\bm{C}_n-\bm{Z}^{(k+1)}), \label{algo-update-lambda}\\
    \bm{\lambda}_2^{(k+1)} &\coloneqq \bm{\lambda}_2^{(k)} + \rho(\bm{q}^{(k+1)} - \bm{S}^{(k+1)}\bm{1}). \label{algo-update-lambda2}
\end{align}

Then, the main algorithm is summarized in Algorithm \ref{algo-LADMM}.
\begin{algorithm}[H]
	\caption{L-ADMM for rLogSpecT}
	\begin{algorithmic}[1]\label{algo-LADMM}
		\REQUIRE Initialization $\bm{S}^{(0)}$, $\bm{\Lambda}^{(0)}$, $\bm{\lambda}_2^{(0)}$, stepsize $\tau$, updating\\
\ \ \ \ \ \   rule for $\rho$, and $k=0$.
		\ENSURE $\bm{S}^{(k)}$ 
		\WHILE{stopping criteria are not satisfied}
		\STATE {Update $\bm{Z}^{(k+1)}$ and $\bm{q}^{(k+1)}$ via \eqref{algo-update-zq};}
		\STATE{Update $\bm{S}^{(k+1)}$ via \eqref{algo-update-S};}
		\STATE{Update $\bm{\Lambda}^{(k+1)}$ via \eqref{algo-update-lambda};}
		\STATE{Update $\bm{\lambda}_2^{(k+1)}$ via \eqref{algo-update-lambda2};}
		\STATE{Update $\rho$ according to the updating rule.}
		\STATE{$k = k+1$;}
		\ENDWHILE
	\end{algorithmic} 
\end{algorithm}
For the convergence analysis of L-ADMM,
% can be directly adapted from \cite[Theorem 4.2]{zhang2011unified}.
we treat $(\bm{Z}, \bm{q})$ as one variable, and the two constraints $\bm{C}_n \bm{S}-\bm{S} \bm{C}_n=\bm{Z}$ and $\bm{q}=\bm{S 1}$ can be written into a single one:
\[
[\bm{I} \otimes \bm{C}_n-\bm{C}_n \otimes \bm{I} ; \bm{I} \otimes \bm{1}^{\top}] \operatorname{vec}(\bm{S})=[\operatorname{vec}(\bm{Z}) ; \bm{q}].
\]
Then we can apply a two-block proximal ADMM (i.e., L-ADMM) to the problem by alternatingly updating $\bm{S}$ (see details in Section \ref{sec-convergence-ana} for this part) and ($\bm{Z}, \bm{q}$). Consequently, the convergence result in \cite[Theorem 4.2]{zhang2011unified} can be invoked directly to derive the following theorem.
\begin{theorem}
    If $\tau > m + \|\bm{C}_n\otimes\bm{I}_m - \bm{I}_m\otimes\bm{C}_n\|^2$, then
    \[\lim_{k \rightarrow \infty}f(\bm{S}^{(k)}) = f_n^*.\]
\end{theorem}

\section{Experiments}
In this section, we present the experiment settings and results on both synthetic and real networks. 

\subsection{Evaluation Metrics}
To evaluate the quality of learned graphs, we use the following standard metrics (see, e.g., \cite{schutze2008introduction}): 
\begin{align*}
	&\text{F-measure}=\frac{2{\sf TP}}{2{\sf TP}+{\sf FN}+{\sf FP}},\\
	&\text{Precision}=\frac{\sf TP}{{\sf TP}+{\sf FP}}, \quad \text{Recall}=\frac{\sf TP}{{\sf TP}+{\sf FN}}.
\end{align*}
Here, ${\sf TP}$, ${\sf FP}$, and ${\sf FN}$ denote the number of true positives, false positives, and false negatives, respectively.
The metrics \emph{precision} and \emph{recall} measure the fraction of correctly retrieved edges among all the edges in the learned graph and in the true graph, respectively. It should be noted that a high value in just one of these two metrics does not imply the graph is accurately learned. This motivates the metric \emph{F-measure}, which is the harmonic mean of precision and recall. A learning algorithm is deemed good if it achieves high F-measure.

\subsection{Data Generation}
\textbf{Random Graphs}. In the experiments, we consider two types of synthetic graphs, namely, the Erd{\H{o}}s-R{\'e}nyi (ER) graph \cite{erdos1960evolution} and the Barabasi-Albert model graph (BA) \cite{barabasi1999emergence}.  The ER graphs are generated by placing an edge between each pair of nodes independently with probability $p=0.2$ and the weight on each edge is set to 1. The BA graphs are generated by having two connected nodes initially and then adding new nodes one at a time, where each new node is connected to exactly one previous node that is randomly chosen with a probability proportional to its degree at the time.

\textbf{Graph Filters}. Three graph filters are used in the experiments. The first one is the low-pass graph filter (lowpass-EXP) $h(\bm{S}) = \exp(\frac{1}{2}\bm{S})$. The second one is the high-pass graph filter (highpass-EXP) $h(\bm{S}) = \exp(-\bm{S})$. The last one is a  quadratic graph filter (QUA) $h(\bm{S}) = \bm{S}^2 + \bm{S} + \bm{I}$. Note that the quadratic graph filter is neither low-pass nor high-pass.

% \begin{figure}[h!]
% 	\centering
% 	\includegraphics[trim=110 30 135 5,clip = true,width=1\linewidth]{fig/performance1211.eps}
% 	\caption{\label{fig:syn-idea}Comparison of LogSpecT and SpecT on Synthetic Data
% 	}
% \end{figure}

\textbf{Graph Signals}. In order to generate stationary graph signals, we follow the generative model in \cite{segarra2017network}:
\[\bm{x} = h(\bm{S})\bm{w},\]
where $\bm{S}$ and $h(\bm{S})$ are respectively the adjacency matrix of the random graph and the graph filter we introduced before, $\bm{w} \sim \mathcal{N}(\bm{0}, \bm{I}_m)$ is a random vector following the normal distribution. As discussed before, $\bm{x}$ is a stationary graph signal and follows a sub-Gaussian distribution.
\subsection{How to Infer Binary Graphs} Many practical tasks require binary graphs instead of weighted graphs. However, the results from LogSpect and rLogSpecT are generally weighted ones. In this section, we tackle the issue of how to convert a weighted graph to a binary one.

Firstly, we normalize each edge in the graph $\bm{W}$ by dividing it over the maximal weight, which yields a collection of values ranging from $0$ to $1$. Our task is then to select a threshold $\varepsilon$ to round these values to $0$ or $1$ and construct a  binary graph $\bm{W}^*$ accordingly. Mathematically, this procedure is captured by the formula:
\begin{equation}\label{infer-binary-graph}
    (W^*)_{ij} = \left\{ \begin{aligned}
        &1 \quad &&\text{if} \ \ W_{ij}/\max\{W_{i'j'}\} \geq \varepsilon, \\
        &0 &&\text{if} \ \ W_{ij}/\max\{W_{i'j'}\} < \varepsilon.
    \end{aligned}
    \right.
\end{equation}
We introduce two schemes to choose the threshold $\varepsilon$: Training-based strategy and searching-based strategy.

\textbf{Training-based}. The training-based strategy trains the model on the training data set with $k$ graphs and chooses $\varepsilon^*$ that can yield the best average performance. For a given graph, we apply $\varepsilon^*$ and follow the formula \eqref{infer-binary-graph} to turn it into a binary one.

\textbf{Searching-based}. This strategy is more straightforward. For a graph, we search for $\varepsilon^*$ that yields the best performance. Compared with training-based strategy, the searching-based strategy is applicable only when the performance of the learned graphs can be measured. Despite its limitation, searching-based strategy generally possesses better performance.
\subsection{Experiments on Synthetic Networks}
\label{exp-syn}
To evaluate the efficacy of LogSpecT and rLogSpecT, we compare them with SpecT and rSpecT on synthetic data. We notice that the prior of no isolated nodes has been incorporated in practice by adding restrictions on the smallest degree. This approach introduces an additional hyperparameter and may exacerbate the infeasibility issue in the robust formulation. Also, no recovery guarantees are developed for this trick. For these reasons, we still compare LogSpect $\&$ rLogSpecT with SpecT $\&$ rSpecT.

\begin{figure}[h!]
	\centering
	\includegraphics[trim=110 30 135 5,clip = true,width=0.8\linewidth]{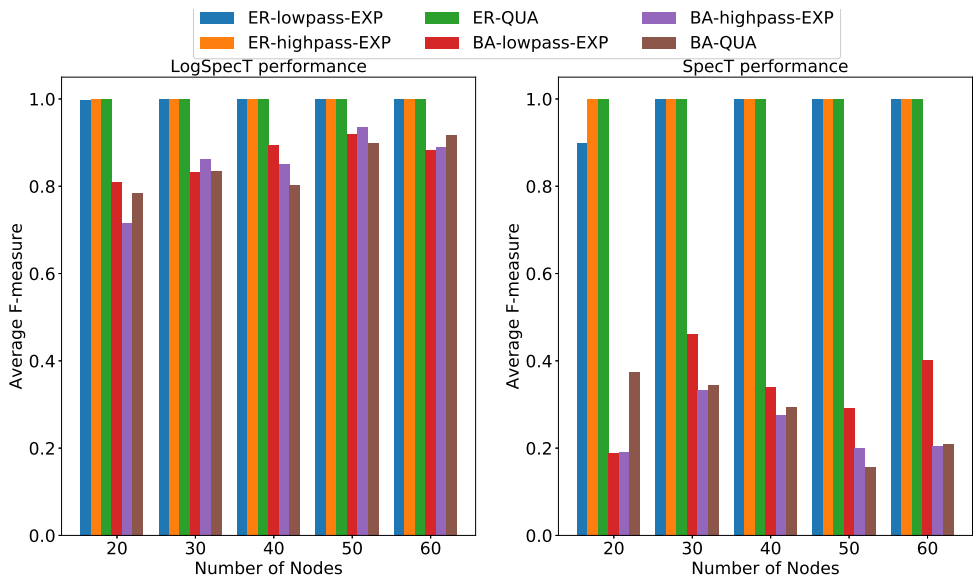}
	\caption{\label{fig:syn-idea}Comparison of LogSpecT and SpecT on Synthetic Data
	}
\end{figure}
\textbf{Performance of LogSpecT}. 
We first compare the performance of the ideal models: LogSpecT and SpecT on the two types of random graphs with three graph filters. We conduct experiments on graphs with nodes ranging from 20 to 60 and use the searching-based strategy to set threshold $\varepsilon^*$. After 10 repetitions, we report the average results in Figure \ref{fig:syn-idea}. The left column presents the performance of LogSpecT and the right column presents that of SpecT. Firstly, we observe that on ER graphs, both models achieve good performance. They nearly recover the graphs perfectly. However, the performance on BA graphs differs. In this case, LogSpecT can work efficiently. The learned graphs from LogSpecT enjoy much higher F-measure values than those from SpecT. Secondly, the comparison between different graph filters shows that different graph filters have few impacts on recovery performance for both LogSpecT and SpecT. Finally, we observe that the outputs of LogSpecT on BA graphs tend to possess higher F-measure when the number of nodes increases. This suggests that LogSpecT may behave better on larger networks. However, such phenomena cannot be observed from SpecT results.

\begin{figure}[h!]
\begin{minipage}[t]{0.49\linewidth}
    \centering		\includegraphics[trim=12 1 56 50,clip = true,width=0.9\textwidth]{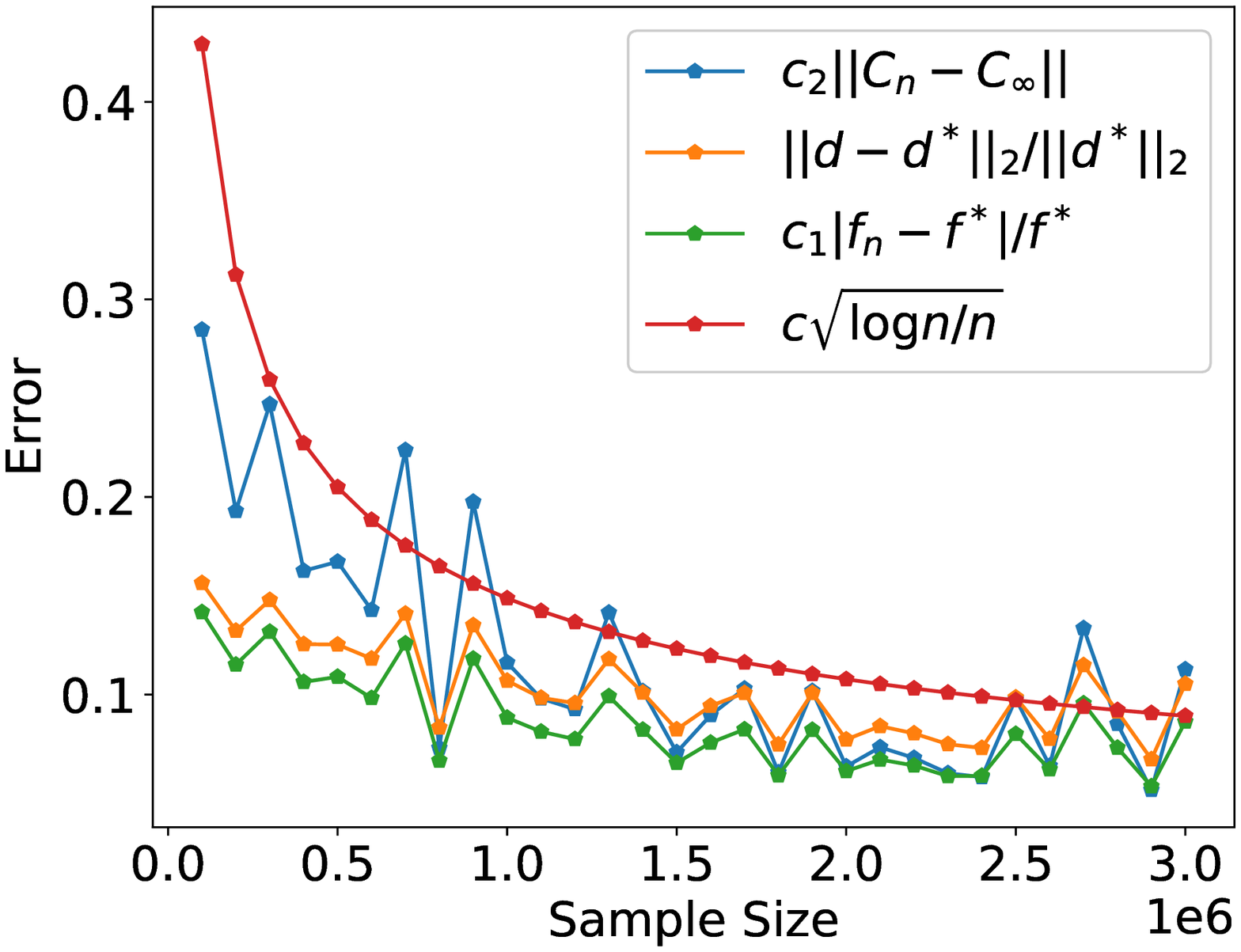}	\caption{\label{fig:syn-samplesize}Trend of Errors for rLogSpecT.}
\end{minipage}
\begin{minipage}[t]{0.49\linewidth}
    \centering		\includegraphics[trim=12 1 56 37,clip = true,width=0.9\textwidth]{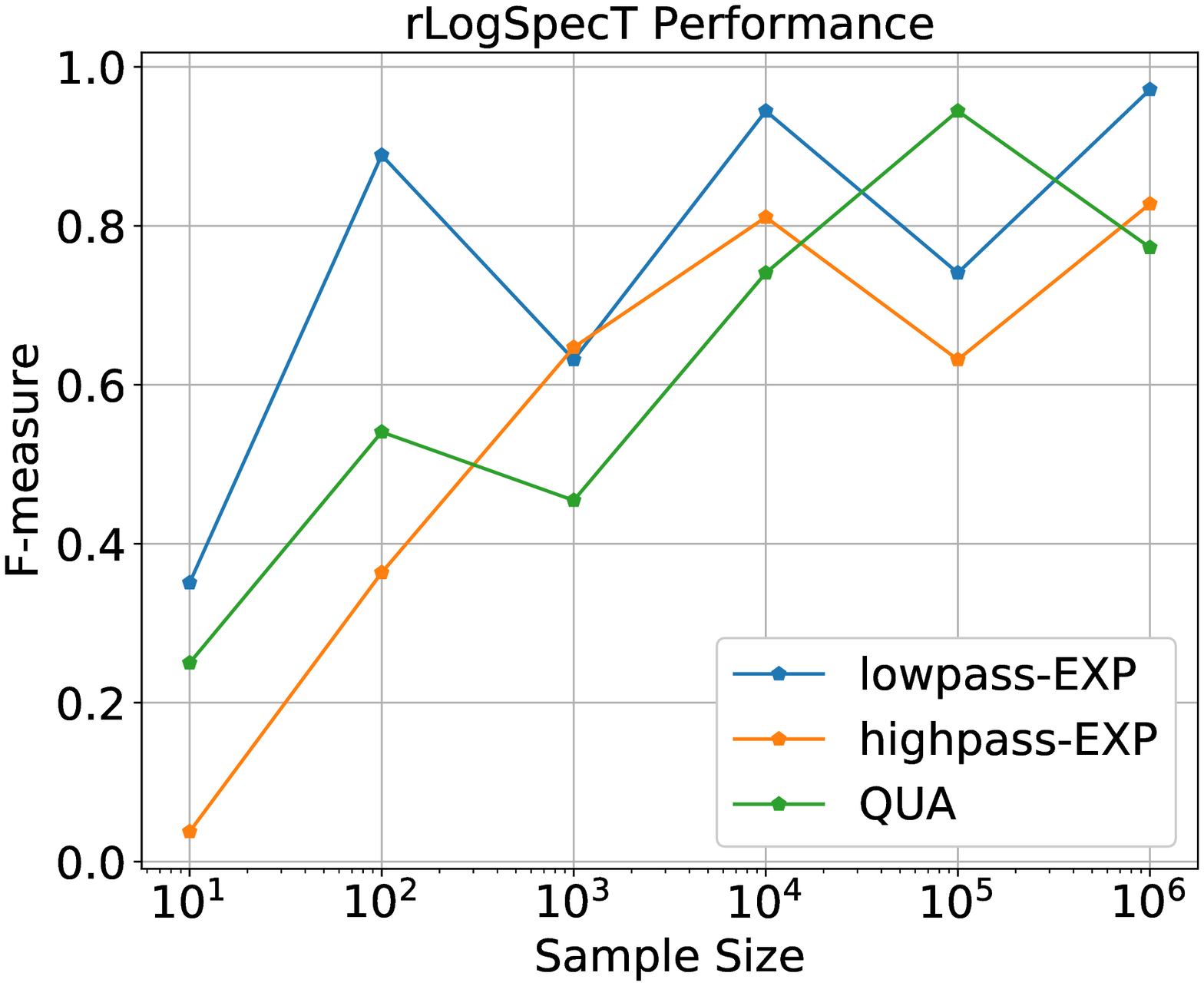}	
    \caption{\label{fig:syn-asymexact}rLogSpecT with $\delta_n = 0.2\sqrt{\log n/n}$.}
\end{minipage}
\end{figure}

\textbf{Performance of rLogSpecT}.
Since LogSpecT persuasively outperforms SpecT, we only present the performance of rLogSpecT when more and more samples are collected. Firstly, we study how the solution to rLogSpecT approaches LogSpecT empirically. To this end, we generate stationary graph signals on ER graphs with 20 nodes from the QUA graph filter. The sample size ranges from $10^5$ to $3*10^5$ and $\delta_n = \|\bm{C}_n - \bm{C}_\infty\|$. We present the experiment results in Figure \ref{fig:syn-samplesize}.
The figure shows that the estimation error $\|\bm{C}_n - \bm{C}\|$, the relative errors of function values $\|f_n - f^*|/f^*$ and of degree vectors $\|\bm{d} - \bm{d}^*\|_2/\|\bm{d}^*\|_2$ are strongly correlated. This indicates that the relative errors between rLogSpecT and LogSpecT in terms of function values and degree vectors can be efficiently controlled by the estimation error of $\bm{C}_\infty$. 

As we have mentioned before, the optimal solution to LogSpecT is not necessarily unique.
% and we can not obtain the non-asymptotic analysis of 
Thus, we do not expect to be able to show
how rLogSpecT's optimal solution converges to LogSpecT's optimal solution. Moreover, the non-uniqueness may jeopardize the performance of rLogSpecT intuitively. The next experiment shows that in practice, the learned graphs from rLogSpecT tend to possess high F-measure values when enough samples are collected. 

rLogSpecT is tested with graph signals from all three graph filters on BA graphs. The sample size $n$ is chosen from $10$ to $10^6$ and $\delta_n$ is set as $0.2\sqrt{\log n/n}$. We rely on the training-based strategy to obtain the best threshold $\varepsilon^*$ from 10 randomly chosen training graphs. We then calculate the F-measure of the learned binary graph in the testing set. The result is reported in Figure \ref{fig:syn-asymexact}. It shows that rLogSpecT works for all three graph filters on BA graphs and tends to achieve better performance when more and more signals are collected. 
%We also observe that compared with unlow-pass (e.g. high-pass and quadratic) filters, performance on low-pass graph filters is marginally better. This leads us to the conjecture that rLogSpecT generally exhibits better performance on low-pass graph filters than others. We will defer the detailed discussions to Appendix \ref{appen-more-synthe}.
This indicates that similar to LogSpecT, rLogSpecT is efficient for different types of graph filters on BA graphs, including the high-pass ones. For more experiments and discussions, we refer readers to Appendix \ref{appen-more-synthe}.

\subsection{Experiments on Real Networks}
% \begin{figure*}[h!]
% \begin{minipage}[t]{0.5\textwidth} 
% \centering
% \includegraphics[trim=65 6 60 34,clip = true,width=\textwidth]{fig/real_ideal.eps}
% \caption{\label{fig:real_ideal}Performance LogSpecT and SpecT.}
% \end{minipage}
%  \begin{minipage}[t]{0.5\textwidth}
% \centering
% \includegraphics[trim=59 6 85 10,clip = true,width=\textwidth]{fig/real_asym.eps}
% \caption{\label{fig:real_asym}Asymptotic performance of rLogSpecT and rSpecT.}
% 	\end{minipage}
% %\caption{\label{fig:real}Performance of different methods on real graphs with synthetic data. }
% \end{figure*}

In this set of experiments, we compare the performance of LogSpecT (resp. rLogSpecT) with SpecT (resp. rSpecT) and other methods from the statistics and GSP communities on a biological graph database \textit{Protein} and a social graph database \textit{Reddit} from \cite{chowdhury2021generalized}. We choose graphs in the databases whose numbers of nodes are smaller than 50. This results in 871 testing graphs in the Protein database and 309 in the Reddit database.

\textbf{Infeasibility of rSpecT}. For these real networks, we first check whether the infeasibility issue encountered by rSpecT is significant. To this end, we adopt the random graph filters $t_1 \bm{S}^2 + t_2\bm{S} + t_3\bm{I}$, where $t_i, i = 1,2,3$ are random variables following a Gaussian distribution with $\mu = 0$ and $\sigma = 2$. Then, we calculate the smallest $\delta_n$ such that rSpecT is feasible. If the smallest $\delta_n > 0$, rSpecT is likely to meet the infeasibility issue. The results with different numbers of graph signals observed are shown in Table \ref{tab:like}.

 \begin{table}[!ht]
    \centering
    \caption{Likelihood of infeasibility \label{tab:like}}
    \begin{tabular}{|c|c|c|c|}
    \hline
        \multicolumn{4}{|c|}{Protein} \\ \hline
        sample size & 10 & 100 & 1000 \\ \hline
        frequency & 0.980 & 0.999 & 0.994 \\ \hline
        mean of $\delta_{\min}$ & 38.514 & 26.012 & 14.813 \\ \hline
        \multicolumn{4}{|c|}{Reddit} \\ \hline
        sample size & 10 & 100 & 1000 \\ \hline
        frequency & 0.984 & 1 & 1 \\ \hline
        mean of $\delta_{\min}$ & 1094.788 & 830.642 & 531.006 \\ \hline
    \end{tabular}
\end{table}
The experiment results indicate a high possibility of encountering infeasibility issues in the real datasets we use, and increasing the sample size does not help to reduce the probability. However, we do observe a decrease in the mean value of $\delta_{\min}$, which makes sense as SpecT should be feasible when the sample size is infinite. The frequent occurrence of infeasibility scenarios necessitates careful consideration when choosing $\delta$ for rSpecT. A commonly used approach is to set $\delta = \delta_{\min}$ \cite{segarra2017network}. However, this method is computationally intensive as it requires an accurate solution to a second-order conic programming (SOCP) in the first stage to obtain $\delta_{\min}$, and it lacks non-asymptotic analysis.

\textbf{Performance of different graph learning models}. The stationary graph signals are generated by the low-pass filter $h(\bm{S}) = \exp(\bm{S})$ on these graphs.
We choose the famous statistical method called (thresholded) correlation \cite{StatisticalNetwork} and the first GSP method that applies the 
 log barrier to graph inference \cite{kalofolias2016learn} as the baselines. The optimal threshold for the correlation method is selected from $0.1$ to $0.6$ and we search in $\{0.01,0.1,1,10,100,1000\}$ to obtain the best hyperparameter in Kalofolias' model. The parameter $\delta_n$ in rLogSpecT is set as $10\sqrt{\log n/n}$ and in rSpecT it is set as the smallest value that allows a feasible solution \cite{segarra2017network}. We also rely on the searching-based strategy to convert the learned graphs from LogSpecT and SpecT to binary ones.

\begin{figure*}[h!]
	\centering		\includegraphics[trim=65 6 60 34,clip = true,width=0.9\textwidth]{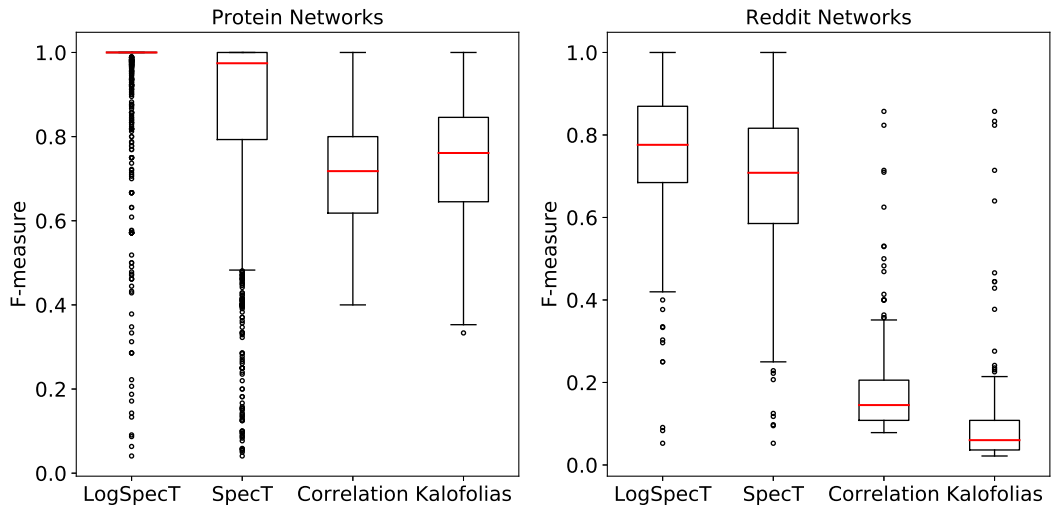}
\caption{\label{fig:real_ideal}Performance LogSpecT and SpecT.}
\end{figure*}

The results of the ideal models with the true covariance matrix $\bm{C}_\infty$ applied are collected in Figure \ref{fig:real_ideal}. We observe that on the real graphs, LogSpecT achieves the best performance on average (median represented by the red line). Also, compared with SpecT, LogSpecT performs more stably. We remark that since the graph signals are not necessarily smooth, Kalofolias' model cannot provide guaranteed performance, especially on the Reddit networks.

\begin{figure*}[h!]
	\centering		\includegraphics[trim=59 6 85 10,clip = true,width=0.9\textwidth]{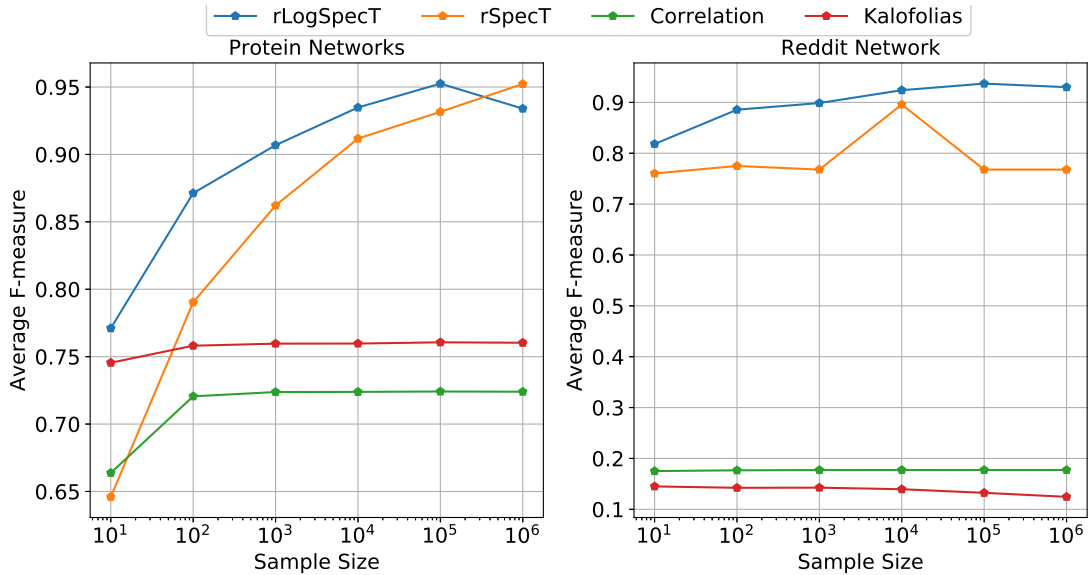}
\caption{\label{fig:real_asym}Asymptotic performance of rLogSpecT and rSpecT.}
\end{figure*}

Figure \ref{fig:real_asym} compares the performance of four different methods when different numbers of signals are observed.\footnote{The model in \cite{shafipour2020online} is a substitute of rSpecT to approximate SpecT and its hyperparmeter is hard to determine. Hence, we omit the performance of that model.} When the sample size increases, the models focusing on stationarity property can recover the graphs more accurately while correlation method and Kalofolias' method fail to exploit the signal information. This can also be inferred from the experiment results in Figure \ref{fig:real_ideal} since the models fail to achieve good performance from full information, let alone from partial information. The experiment also shows that when a fixed number of samples are observed, the learned graphs from rLogSpecT approximate the ground truth better than rSpecT. This further corroborates the superiority of rLogSpecT on graph learning from stationary signals.

\section{Conclusion}
In this paper, we provide the first infeasibility condition for the most fundamental model of graph learning from stationary signals \cite{segarra2017network} and propose an efficient alternative. The recovery guarantees of the robust formulation for the finite-sample case are then analyzed with advanced optimization tools, which may find broader applications in learning tasks. Compared with current literature \cite{navarro2020joint}, these theoretical results require less stringent conditions. We also design an L-ADMM algorithm that allows for efficient implementation and theoretical convergence. Extensive experiments on both synthetic and real data are conducted in this paper. The results show that our proposed model can significantly outperform existing ones (e.g. experiments on BA graphs).

We believe this work represents an important step beyond the fundamental model. Its general formulation allows for the transfer of the current extensions of SpecT. Testing our proposed models with these extensions is one future direction. Also, we notice that although the recovery guarantees for robust formulations are clear, the estimation performance analysis for the ideal-case models (i.e. SpecT and LogSpecT) is still incomplete. Investigating the exact recovery conditions is another future direction.

\nocite{}
\bibliography{arXiv}

\newcommand{\etalchar}[1]{$^{#1}$}
\begin{thebibliography}{dMCYP21}

\bibitem[AOTS15]{acemoglu2015systemic}
Daron Acemoglu, Asuman Ozdaglar, and Alireza Tahbaz-Salehi.
\newblock Systemic risk and stability in financial networks.
\newblock {\em American Economic Review}, 105(2):564--608, 2015.

\bibitem[BA99]{barabasi1999emergence}
Albert-L{\'a}szl{\'o} Barab{\'a}si and R{\'e}ka Albert.
\newblock Emergence of scaling in random networks.
\newblock {\em science}, 286(5439):509--512, 1999.

\bibitem[BPC{\etalchar{+}}11]{boyd2011distributed}
Stephen Boyd, Neal Parikh, Eric Chu, Borja Peleato, Jonathan Eckstein, et~al.
\newblock Distributed optimization and statistical learning via the alternating
  direction method of multipliers.
\newblock {\em Foundations and Trends{\textregistered} in Machine learning},
  3(1):1--122, 2011.

\bibitem[BRCM19]{buciulea2019network}
Andrei Buciulea, Samuel Rey, Cristobal Cabrera, and Antonio~G Marques.
\newblock Network reconstruction from graph-stationary signals with hidden
  variables.
\newblock In {\em Proceedings of the 53rd Asilomar Conference on Signals,
  Systems, and Computers}, pages 56--60. IEEE, 2019.

\bibitem[BRM22]{buciulea2022learning}
Andrei Buciulea, Samuel Rey, and Antonio~G Marques.
\newblock Learning graphs from smooth and graph-stationary signals with hidden
  variables.
\newblock {\em IEEE Transactions on Signal and Information Processing over
  Networks}, 8:273--287, 2022.

\bibitem[CN21]{chowdhury2021generalized}
Samir Chowdhury and Tom Needham.
\newblock Generalized spectral clustering via gromov-wasserstein learning.
\newblock In {\em Proceedings of the 24th International Conference on
  Artificial Intelligence and Statistics (AISTATS 2021)}, pages 712--720. PMLR,
  2021.

\bibitem[dMCYP21]{de2021graphical}
Jose~Vinicius de~Miranda~Cardoso, Jiaxi Ying, and Daniel Palomar.
\newblock Graphical models in heavy-tailed markets.
\newblock {\em Advances in Neural Information Processing Systems},
  34:19989--20001, 2021.

\bibitem[ER{\etalchar{+}}60]{erdos1960evolution}
Paul Erd{\H{o}}s, Alfr{\'e}d R{\'e}nyi, et~al.
\newblock On the evolution of random graphs.
\newblock {\em Publication of the Mathematical Institute of the Hungarian
  Academy of Sciences}, 5(1):17--60, 1960.

\bibitem[Gir15]{girault2015stationary}
Benjamin Girault.
\newblock Stationary graph signals using an isometric graph translation.
\newblock In {\em Proceedings of the 23rd European Signal Processing Conference
  (EUSIPCO 2015)}, pages 1516--1520. IEEE, 2015.

\bibitem[Hof52]{hoffman1952approximate}
Alan~J Hoffman.
\newblock On approximate solutions of systems of linear inequalities.
\newblock {\em Journal of Research of the National Bureau of Standards},
  49(4):263--265, 1952.

\bibitem[HW22]{he2022detecting}
Yiran He and Hoi-To Wai.
\newblock Detecting central nodes from low-rank excited graph signals via
  structured factor analysis.
\newblock {\em IEEE Transactions on Signal Processing}, 70:2416--2430, 2022.

\bibitem[JHIM{\etalchar{+}}19]{jung2019semi}
Alexander Jung, Alfred~O Hero~III, Alexandru~Cristian Mara, Saeed Jahromi,
  Ayelet Heimowitz, and Yonina~C Eldar.
\newblock Semi-supervised learning in network-structured data via total
  variation minimization.
\newblock {\em IEEE Transactions on Signal Processing}, 67(24):6256--6269,
  2019.

\bibitem[JT19]{jung2019localized}
Alexander Jung and Nguyen Tran.
\newblock Localized linear regression in networked data.
\newblock {\em IEEE Signal Processing Letters}, 26(7):1090--1094, 2019.

\bibitem[Kal16]{kalofolias2016learn}
Vassilis Kalofolias.
\newblock How to learn a graph from smooth signals.
\newblock In {\em Proceedings of the 19th International Conference on
  Artificial Intelligence and Statistics (AISTATS 2016)}, pages 920--929. PMLR,
  2016.

\bibitem[KLTF17]{kalofolias2017learning}
Vassilis Kalofolias, Andreas Loukas, Dorina Thanou, and Pascal Frossard.
\newblock Learning time varying graphs.
\newblock In {\em Proceedings of 2017 IEEE International Conference on
  Acoustics, Speech and Signal Processing (ICASSP 2017)}, pages 2826--2830.
  IEEE, 2017.

\bibitem[Kol09]{StatisticalNetwork}
Eric~D. Kolaczyk.
\newblock {\em Statistical Analysis of Network Data: Methods and Models}.
\newblock Springer-Verlag, New York, NY, USA, 2009.

\bibitem[MF20]{maretic2020graph}
Hermina~Petric Maretic and Pascal Frossard.
\newblock Graph laplacian mixture model.
\newblock {\em IEEE Transactions on Signal and Information Processing over
  Networks}, 6:261--270, 2020.

\bibitem[MNBD21]{marti2021review}
Gautier Marti, Frank Nielsen, Miko{\l}aj Bi{\'n}kowski, and Philippe Donnat.
\newblock A review of two decades of correlations, hierarchies, networks and
  clustering in financial markets.
\newblock {\em Progress in Information Geometry}, pages 245--274, 2021.

\bibitem[MSLR17]{marques2017stationary}
Antonio~G Marques, Santiago Segarra, Geert Leus, and Alejandro Ribeiro.
\newblock Stationary graph processes and spectral estimation.
\newblock {\em IEEE Transactions on Signal Processing}, 65(22):5911--5926,
  2017.

\bibitem[MSMR19]{mateos2019connecting}
Gonzalo Mateos, Santiago Segarra, Antonio~G Marques, and Alejandro Ribeiro.
\newblock Connecting the dots: Identifying network structure via graph signal
  processing.
\newblock {\em IEEE Signal Processing Magazine}, 36(3):16--43, 2019.

\bibitem[MSR08]{schutze2008introduction}
D.~Christopher Manning, Hinrich Sch{\"u}tze, and Prabhakar Raghavan.
\newblock {\em Introduction to Information Retrieval}.
\newblock Cambridge University Press, 2008.

\bibitem[NWM{\etalchar{+}}22]{navarro2020joint}
Madeline Navarro, Yuhao Wang, Antonio~G Marques, Caroline Uhler, and Santiago
  Segarra.
\newblock Joint inference of multiple graphs from matrix polynomials.
\newblock {\em Journal of Machine Learning Research}, 23(76):1--35, 2022.

\bibitem[PGM{\etalchar{+}}17]{pasdeloup2017characterization}
Bastien Pasdeloup, Vincent Gripon, Gr{\'e}goire Mercier, Dominique Pastor, and
  Michael~G Rabbat.
\newblock Characterization and inference of graph diffusion processes from
  observations of stationary signals.
\newblock {\em IEEE transactions on Signal and Information Processing over
  Networks}, 4(3):481--496, 2017.

\bibitem[PV17]{perraudin2017stationary}
Nathana{\"e}l Perraudin and Pierre Vandergheynst.
\newblock Stationary signal processing on graphs.
\newblock {\em IEEE Transactions on Signal Processing}, 65(13):3462--3477,
  2017.

\bibitem[RBN{\etalchar{+}}22]{rey2022joint}
Samuel Rey, Andrei Buciulea, Madeline Navarro, Santiago Segarra, and Antonio~G
  Marques.
\newblock Joint inference of multiple graphs with hidden variables from
  stationary graph signals.
\newblock In {\em Proceedings of 2022 IEEE International Conference on
  Acoustics, Speech and Signal Processing (ICASSP 2022)}, pages 5817--5821.
  IEEE, 2022.

\bibitem[RW09]{rockafellar2009variational}
R~Tyrrell Rockafellar and Roger J-B Wets.
\newblock {\em Variational analysis}, volume 317.
\newblock Springer Science \& Business Media, 2009.

\bibitem[RW21]{royset2021optimization}
Johannes~O Royset and Roger~JB Wets.
\newblock {\em An Optimization Primer}.
\newblock Springer, 2021.

\bibitem[RWS20]{ramakrishna2020user}
Raksha Ramakrishna, Hoi-To Wai, and Anna Scaglione.
\newblock A user guide to low-pass graph signal processing and its
  applications: Tools and applications.
\newblock {\em IEEE Signal Processing Magazine}, 37(6):74--85, 2020.

\bibitem[SCML18]{segarra2018statistical}
Santiago Segarra, Sundeep~Prabhakar Chepuri, Antonio~G Marques, and Geert Leus.
\newblock Statistical graph signal processing: Stationarity and spectral
  estimation.
\newblock {\em Cooperative and Graph Signal Processing}, pages 325--347, 2018.

\bibitem[SHMV19]{shafipour2019online}
Rasoul Shafipour, Abolfazl Hashemi, Gonzalo Mateos, and Haris Vikalo.
\newblock Online topology inference from streaming stationary graph signals.
\newblock In {\em 2019 IEEE Data Science Workshop (DSW)}, pages 140--144. IEEE,
  2019.

\bibitem[SM20]{shafipour2020online}
Rasoul Shafipour and Gonzalo Mateos.
\newblock Online topology inference from streaming stationary graph signals
  with partial connectivity information.
\newblock {\em Algorithms}, 13(9):228, 2020.

\bibitem[SMMR17]{segarra2017network}
Santiago Segarra, Antonio~G Marques, Gonzalo Mateos, and Alejandro Ribeiro.
\newblock Network topology inference from spectral templates.
\newblock {\em IEEE Transactions on Signal and Information Processing over
  Networks}, 3(3):467--483, 2017.

\bibitem[SMSK{\etalchar{+}}11]{smith2011network}
Stephen~M Smith, Karla~L Miller, Gholamreza Salimi-Khorshidi, Matthew Webster,
  Christian~F Beckmann, Thomas~E Nichols, Joseph~D Ramsey, and Mark~W Woolrich.
\newblock Network modelling methods for fmri.
\newblock {\em Neuroimage}, 54(2):875--891, 2011.

\bibitem[STM15]{stegle2015computational}
Oliver Stegle, Sarah~A Teichmann, and John~C Marioni.
\newblock Computational and analytical challenges in single-cell
  transcriptomics.
\newblock {\em Nature Reviews Genetics}, 16(3):133--145, 2015.

\bibitem[SWUM17]{segarra2017joint}
Santiago Segarra, Yuhao Wang, Caroline Uhler, and Antonio~G Marques.
\newblock Joint inference of networks from stationary graph signals.
\newblock In {\em Proceedings of the 51st Asilomar Conference on Signals,
  Systems, and Computers}, pages 975--979. IEEE, 2017.

\bibitem[TE20]{tanaka2020generalized}
Yuichi Tanaka and Yonina~C Eldar.
\newblock Generalized sampling on graphs with subspace and smoothness priors.
\newblock {\em IEEE Transactions on Signal Processing}, 68:2272--2286, 2020.

\bibitem[TEOC20]{tanaka2020sampling}
Yuichi Tanaka, Yonina~C Eldar, Antonio Ortega, and Gene Cheung.
\newblock Sampling signals on graphs: From theory to applications.
\newblock {\em IEEE Signal Processing Magazine}, 37(6):14--30, 2020.

\bibitem[Ver12]{vershynin2012close}
Roman Vershynin.
\newblock How close is the sample covariance matrix to the actual covariance
  matrix?
\newblock {\em Journal of Theoretical Probability}, 25(3):655--686, 2012.

\bibitem[Ver18]{vershynin2018high}
Roman Vershynin.
\newblock {\em High-dimensional probability: An introduction with applications
  in data science}, volume~47.
\newblock Cambridge University Press, 2018.

\bibitem[WYLS21]{wang2021efficient}
Xiaolu Wang, Chaorui Yao, Haoyu Lei, and Anthony Man-Cho So.
\newblock An efficient alternating direction method for graph learning from
  smooth signals.
\newblock In {\em Proceedings of 2021 IEEE International Conference on
  Acoustics, Speech and Signal Processing (ICASSP 2021)}, pages 5380--5384.
  IEEE, 2021.

\bibitem[ZBO11]{zhang2011unified}
Xiaoqun Zhang, Martin Burger, and Stanley Osher.
\newblock A unified primal-dual algorithm framework based on bregman iteration.
\newblock {\em Journal of Scientific Computing}, 46(1):20--46, 2011.

\bibitem[ZJSL17]{zhu2017learning}
Zhaowei Zhu, Shengda Jin, Xuming Song, and Xiliang Luo.
\newblock Learning graph structure with stationary graph signals via
  first-order approximation.
\newblock In {\em Proceedings of the 22nd International Conference on Digital
  Signal Processing (DSP 2017)}, pages 1--5. IEEE, 2017.

\bibitem[ZWKP19]{zhao2019admmforgraph}
Licheng Zhao, Yiwei Wang, Sandeep Kumar, and Daniel~P Palomar.
\newblock Optimization algorithms for graph laplacian estimation via admm and
  mm.
\newblock {\em IEEE Transactions on Signal Processing}, 67(16):4231--4244,
  2019.

\bibitem[ZYY16]{zhang2016one}
Hui Zhang, Ming Yan, and Wotao Yin.
\newblock One condition for solution uniqueness and robustness of both
  l1-synthesis and l1-analysis minimizations.
\newblock {\em Advances in Computational Mathematics}, 42(6):1381--1399, 2016.

\bibitem[ZZL{\etalchar{+}}22]{zheng2022multi}
Shuai Zheng, Zhenfeng Zhu, Zhizhe Liu, Zhenyu Guo, Yang Liu, Yuchen Yang, and
  Yao Zhao.
\newblock Multi-modal graph learning for disease prediction.
\newblock {\em IEEE Transactions on Medical Imaging}, 41(9):2207--2216, 2022.

\end{thebibliography}
\bibliographystyle{alpha}

\newpage
\section*{\LARGE Appendix}
\begin{appendices}
% \section{Proof of Lemma \ref{lemma-exp-lip}, \ref{lemma-dist-rotation} and \ref{lemma-inj-cvx-radius} 
% \appendix
% \section{Proofs in Section \ref{section-infea}.}
% This section contains the proofs for results in section \ref{section-infea}.
The appendix includes the missing proofs, detailed discussions of some argument in the main body and more numerical experiments.
\section{Proof of Theorem \ref{thm-infeasibility}}
% \begin{proof}
     Since the linear system \eqref{thm-infea-linear} has no solution, we know from Farkas' lemma that the following system has solutions:
    % \begin{equation}\label{thm-infea-linear21}
    %     \begin{split}
    %         \left[ \bm{I}_{m-1}\quad \bm{0}_{\frac{(m-1)(m-2)}{2}} \right]\bm{B}^\top\bm{A}^\top\bm{x} &\leq \bm{-1}_{(m-1)\times 1}, \\
    %         \left[ \bm{0}_{\frac{(m-1)(m-2)}{2}\times (m-1)}\quad \bm{I}_{\frac{(m-1)(m-2)}{2}} \right]\bm{B}^\top\bm{A}^\top\bm{x} &\leq \bm{0}_{\frac{(m-1)(m-2)}{2}\times 1},
    %     \end{split}
    % \end{equation}
    % which is equivalent to:
    \begin{equation}\label{thm-infea-linear2}
       \left\{ \begin{split}
            \left[ \bm{I}_{m-1}\quad \bm{0}_{\frac{(m-1)(m-2)}{2}} \right]\bm{B}^\top\bm{A}_n^\top\bm{x} &< \bm{0}_{(m-1)\times 1}, \\
            \left[ \bm{0}_{\frac{(m-1)(m-2)}{2}\times (m-1)}\quad \bm{I}_{\frac{(m-1)(m-2)}{2}} \right]\bm{B}^\top\bm{A}_n^\top\bm{x} &\leq \bm{0}_{\frac{(m-1)(m-2)}{2}\times 1}.
        \end{split}
        \right.
    \end{equation}
Let $\bm{x}^*\in\mathbb{R}^{m^2}$ be a solution to \eqref{thm-infea-linear2}. Denote $\bm{x}_+ \coloneqq \max\{\bm{x}^*,\bm{0}\}$, $\bm{x}_- \coloneqq \max\{-\bm{x}^*,\bm{0}\}$. 
    % and $\bm{b}\in \mathbb{R}^{\frac{1}{2}m(m-1)}$ satisfying 
    % \begin{equation*}
    %     b_i = \left\{\begin{aligned}
    %      1,\ \ &\text{if}\quad i \in [m-1] \\
    %      0,\ \ &\text{if} \quad i >m-1
    %     \end{aligned}
    %     \right.
    % \end{equation*}
    Then, there exists $c \in (0,1]$ such that 
    \begin{equation*}
        \bm{B}^\top\bm{A}_n^\top(\bm{x}_+ -\bm{x}_-)+c\bm{1}_{m^2}^\top(\bm{x}_+ +\bm{x}_-)[\bm{1}_{m-1}; \bm{0}_{\frac{(m-1)(m-2)}{2}}] \leq \bm{0}.
    \end{equation*}
    Define $y \coloneqq -\bm{1}_{m^2}^\top(\bm{x}_+ +\bm{x}_-)$, $z \coloneqq c\bm{1}_{m^2}^\top(\bm{x}_+ +\bm{x}_-)$ 
    % \begin{align*}
    % {x}_3 &= -\bm{1}^\top(\bm{x}_1+\bm{x}_2), \\
    %     {x}_4 &= k\bm{1}^\top(\bm{x}_1+\bm{x}_2),
    % \end{align*}
    and set $\bar{\delta} = c$. For all $\delta \in [0, \bar{\delta})$, $(\bm{x}_+,\bm{x}_-,y,z)$ is a solution to the following linear system:
    \begin{equation*}
    \left\{
        \begin{split}
            \bm{B}^\top\bm{A}_n^\top(\bm{x}_+ -\bm{x}_-) + z[\bm{1}_{m-1}; \bm{0}_{\frac{(m-1)(m-2)}{2}}] &\leq \bm{0}, \\
            \bm{1}_{m^2}^\top(\bm{x}_+ +\bm{x}_-) + y &\leq 0, \\ 
            \delta y + z &> 0, \\
            \bm{x}_+, \bm{x}_-, -y &\geq \bm{0}.
        \end{split}
        \right.
    \end{equation*}
    Again, from Farkas' lemma, this implies that the following linear system does not have a solution:
    \begin{equation}\label{thm-infea-linear3}
    \left\{
        \begin{split}
            \bm{A}_n\bm{B}\bm{s}+t\bm{1}_{m^2} &\geq \bm{0}, \\
            \bm{A}_n\bm{B}\bm{s}-t\bm{1}_{m^2} &\leq \bm{0},\\
            t &\leq \delta,\\
            \left[\bm{1}_{m-1}\quad \bm{0}_{\frac{(m-1)(m-2)}{2}}\right]\bm{s} &= 1,
        \end{split}
        \right.
    \end{equation}
    where $\bm{s}\in \mathbb{R}^{m(m-1)/2}$ and $t\in \mathbb{R}$. Since \eqref{thm-infea-linear3} is equivalent to:
    \begin{equation}\label{thm-infea-linear4}
    \left\{
        \begin{split}
            \|\bm{C}_n\bm{S}-\bm{S}\bm{C}_n\|_{\infty,\infty} \leq \delta, \\
            (\bm{S}\bm{1})_1 = 1, \\
            \bm{S}\in \mathcal{S},
        \end{split}
        \right.
    \end{equation}
    the above argument indicates that \eqref{thm-infea-linear4} does not have a solution. 
    % Next, we prove the theorem by contradiction. 
    Suppose rSpecT has a feasible solution $\bm{S}'$, then 
    \begin{equation*}
        \|\bm{C}_n\bm{S}'-\bm{S}'\bm{C}_n\|_{\infty,\infty} \leq  \|\bm{C}_n\bm{S}'-\bm{S}'\bm{C}_n\|_F \leq \delta.
    \end{equation*}
    % \Lingzhi{Does it hold?}
    Hence, $\bm{S}'$ is also a solution to \eqref{thm-infea-linear4}. However, \eqref{thm-infea-linear4} does not have a solution. We can conclude that rSpecT is infeasible in this case.
% \end{proof}

\section{Explanations on Sufficient Conditions in Theorem \ref{thm-infeasibility}}
We elaborate more on the infeasibility condition that$\bm{A}_n\bm{B}$ has full column rank. An application of the condition is Example \ref{exm-infeas-2nodes}. Specifically, we know that in this case,
% in this case
\begin{align*}
    &\bm{B} = \left(\begin{aligned}
        &0 \\
        &1 \\
        &1 \\
        &0
    \end{aligned}\right)\quad\text{and}\quad \bm{A}_n = \begin{pmatrix}
        0 & h_{12} & -h_{12} & 0 \\
        h_{12} & h_{22} - h_{11} & 0 & -h_{12} \\
        -h_{12} & 0 & h_{11} - h_{22} & h_{12} \\
        0 & -h_{12} & h_{12} & 0
    \end{pmatrix}. 
\end{align*}
\noindent \text{This implies that}  
\begin{align*}
  &\bm{A}_n\bm{B} = \begin{pmatrix}
     0 \\
     h_{22} - h_{11} \\
     h_{11} - h_{22} \\
     0
 \end{pmatrix}.
\end{align*}
Hence, when $h_{11} \neq h_{22}$, $\bm{A}_n\bm{B}$ has full column rank. This means that when $\delta$ is small enough (from Example \ref{exm-infeas-2nodes} we know $\Tilde{\delta} = \sqrt{2}|h_{11} - h_{22}|$), the model rSpecT is infeasible.
\section{Proofs of Properties of (r)LogSpecT
% Section \ref{section-recovery}
}
\subsection{Proof of Proposition \ref{theorem-choice-parameters}}
% \begin{proof}
        % The following proof relies on the fact that 
        Since the constraint set $\mathcal{S}$ is a cone, it follows that for all $\gamma > 0$, $\gamma \mathcal{S} = \mathcal{S}$. Then, we know that 
    \begin{equation*}
    \begin{aligned}
        \Opt(\bm{C}, \alpha) &= \underset{\bm{S}\in \mathcal{S}, \bm{C}\bm{S} = \bm{S}\bm{C}}{\argmin} \ \|\bm{S}\|_{1,1} - \alpha\bm{1}^\top\log(\bm{S}\bm{1}) 
         \\
        &= \gamma \cdot  \underset{\gamma\bm{S}\in \mathcal{S}, \bm{C}\gamma\bm{S} = \gamma\bm{S}\bm{C}}{\argmin }\ \|\gamma\bm{S}\|_{1,1} - \alpha\bm{1}^\top\log(\gamma\bm{S}\bm{1}) \\
        &= \gamma\cdot \underset{\bm{S}\in \frac{1}{\gamma}\mathcal{S}, \bm{C}\bm{S} = \bm{S}\bm{C}}{\argmin}\ \gamma\|\bm{S}\|_{1,1} - \alpha\bm{1}^\top\log(\bm{S}\bm{1})  \\
        &=\gamma\cdot \underset{\bm{S}\in \mathcal{S}, \bm{C}\bm{S} = \bm{S}\bm{C}}{\argmin}\  \|\bm{S}\|_{1,1} - \frac{\alpha}{\gamma}\bm{1}^\top\log(\bm{S}\bm{1})  \\
        &= \gamma \Opt(\bm{C},\alpha/\gamma),
    \end{aligned}
    \end{equation*}
    where the third equality is from the basic calculus rule of the logarithm function.
    Set $\gamma = \alpha$ and then $\Opt(\bm{C}, \alpha)=\alpha\Opt(\bm{C}, 1)$, which completes the proof.
% \end{proof}

\subsection{Proof of Proposition \ref{prop-feasible}}
% \begin{proof}
The proof will be conducted by constructing a feasible solution for rLogSpecT. Recall that $\bm{A}_n = \bm{I} \otimes \bm{C}_n - \bm{C}_n \otimes \bm{I}$ and the matrix $\bm{B}\in \mathbb{R}^{m^2\times m(m-1)/2}$ that maps a non-negative vector  to the vectorization of a valid adjacency matrix. Let $\bm{S}=\min\{\frac{\delta}{\|\bm{A}_n\bm{B}\bm{s}\|_2}, 1\}\cdot\text{mat}(\bm{B}\bm{s})$  with $\bm{s} \in \mathbb{R}^{(m-1)m/2}$ being a non-negative vector, where mat$(\cdot)$ is the matricization operator. Note that 
\begin{equation*}
	\text{vec}(\bm{C}_n\bm{S}-\bm{S}\bm{C}_n) = \left(\bm{I} \otimes \bm{C}_n - \bm{C}_n \otimes \bm{I}\right)\text{vec}(\bm{S})=\bm{A}_n\text{vec}(\bm{S}).
\end{equation*}
Then, we know that
\begin{align*}
    \|\bm{C}_n\bm{S} - \bm{S}\bm{C}_n\|_F 
     = \|\text{vec}(\bm{C}_n\bm{S}-\bm{S}\bm{C}_n)\|_2 = \min\left\{\frac{\delta}{\|\bm{A}_n\bm{B}\bm{s}\|_2}, 1\right\}\cdot\|\bm{A}_n\bm{B}\bm{s}\|_2 \leq \delta.
\end{align*}
% where $\hat{\bm{S}}$ is the matrix form of $\bm{A}\bm{B}\hat{\bm{s}}$. 
Thus, the given $\bm{S}$ is a feasible solution for rLogSpecT and it completes the proof.
% Since $\bm{s}$ can induce a matrix in the domain of the objective, $\hat{\bm{S}}$ is also in the domain of the objective. Hence, we get a feasible solution to rLogSpecT. This proves the proposition.

% We denote $\bm{A} \coloneqq \bm{I} \otimes \bm{C} - \bm{C} \otimes \bm{I}$. 
% \end{proof}

% This section provides proof for the recovery guarantee of rLogSpecT.
\subsection{Proof of Proposition \ref{lemma-bound-solutions}}
% \begin{proof}
For the first statement, let us consider the Karush-Kuhn-Tucker (KKT) conditions of LogSpecT and rLogSpecT. Since the LogSpecT is a convex problem and Slater's condition holds, the KKT conditions are necessary and sufficient for the optimality,
% . From the KKT conditions of problem \eqref{model-novel-ideal} we know that 
i.e., there exists $(\bm{\Lambda}_1, \bm{\Lambda}_2) \in \mathbb{R}^{m\times m}\times \mathcal{N}_{\mathcal{S}}(\bm{S}^*)$ such that
    \begin{equation}\label{equa-optcon-ideal}
    \left\{
    \begin{aligned}
        % &\exists (\bm{\Lambda}_1, \bm{\Lambda}_2) \in \mathbb{R}^{m\times m}\times \mathcal{N}_{\mathcal{S}}(\bm{S}^*), \text{ such that} \\
        &\nabla_{\bm{S}} (\|\bm{S}^*\|_{1,1} - \alpha  \bm{1}^\top\log(\bm{S}^*\bm{1})) + \bm{C}_\infty\bm{\Lambda}_1 - \bm{\Lambda}_1\bm{C}_\infty + \bm{\Lambda}_2 = \bm{0}, \\
        &\bm{C}_\infty\bm{S}^* = \bm{S}^*\bm{C}_\infty,\\ 
        &\bm{S}^* \in \mathcal{S},
    \end{aligned}
    \right.
    \end{equation}
    where $\mathcal{N}_{\mathcal{S}}(\bm{S}^*)\coloneqq\{\bm{N}\in\mathbb{R}^{m\times m}: \sup_{\bm{X}\in \mathcal{S}}\, \langle \bm{X}-\bm{S}^*, \bm{N}\rangle \leq0\}$ is the normal cone of $\mathcal{S}$ at $\bm{S}^*$, and $\nabla \|\bm{S}^*\|_{1,1}$ is well-defined since $\|\cdot\|_{1,1} = \langle \cdot,\bm{1}\bm{1}^\top \rangle$ at $\bm{S}^*\ge 0$, which is differentiable.  Taking further calculation gives that 
    % since $\bm{S}\ge 0$, $\|\cdot\|_{1,1}$ is differentiable
	\begin{align*}
		\nabla \|\bm{S}^*\|_{1,1}=\bm{11}^\top,\quad (\nabla_{\bm{S}} \bm{1}^\top\log(\bm{S}^*\bm{1}))_{ij} = \frac{1}{(\bm{S}^*\bm{1})_i}.
	\end{align*}
	Combining this with \eqref{equa-optcon-ideal} by taking inner product of both sides  with $\bm{S}^*$, we obtain that
	\begin{align}\label{equa-optcon-ideal-2}
		\sum_{i,j}(\bm{S}^*)_{ij} - \alpha\sum_{i,j}\frac{(\bm{S}^*)_{ij}}{(\bm{S}^* \bm{1})_i}+ \langle\bm{\Lambda}_1, \bm{C}_\infty\bm{S}^* - \bm{S}^*\bm{C}_\infty\rangle  + \langle \bm{\Lambda}_2, \bm{S}^* \rangle = 0.
	\end{align}
 % where $d_i \coloneqq $.
	From the structure of $\mathcal{S}$ and the fact that $\Lambda_2\in \mathcal{N}_{\mathcal{S}}(\bm{S}^*)$, one has that $\langle \bm{\Lambda}_2, \bm{S}^* \rangle = 0$. Also, note that $\bm{C}_\infty\bm{S}^* = \bm{S}^*\bm{C}_\infty$. Hence, the equation \eqref{equa-optcon-ideal-2} can be simplified as the desired result:
	\begin{align*}
	    \|\bm{S}^*\|_{1,1}=\sum_{i,j}(\bm{S}^*)_{ij} = \alpha\sum_{i,j}\frac{(\bm{S}^*)_{ij}}{(\bm{S}^*\bm{1})_i} = \alpha\sum_{i = 1}^m\sum_{j = 1}^m \frac{(\bm{S}^*)_{ij}}{(\bm{S}^*\bm{1})_i} = \alpha m.
	\end{align*}
	% Since $\bm{S}^*\in \mathcal{S}$, $\bm{S}^*_{ij}\geq 0$. The above relation equals to 
	% \begin{equation*}
	%     \|\bm{S}^*\|_{1,1} = \alpha m.
	% \end{equation*}
	% This completes the proof of the first part of the lemma.
	
	 The KKT conditions of  rLogSpecT indicate that there exist $\lambda_1\ge0$, $\bm{\Lambda}_2 \in \mathcal{N}_{\mathcal{S}}(\bm{S}_n^*)$
 % , which is the normal cone of $\mathcal{S}$ at $\bm{S}_n^*$, 
 and $\bm{Q} \in \partial \|\bm{C}_n\bm{S}_n^*-\bm{S}_n^*\bm{C}_n\|_F$ (i.e., the subgradient of the function $\bm{S}\mapsto\|\bm{C}_n\bm{S} - \bm{S}\bm{C}_n\|_F$ at $\bm{S}_n^*$) such that
    \begin{equation}\label{equa-optcon-novel}
    \left\{
    \begin{aligned}
		&\nabla_{\bm{S}} (\|\bm{S}_n^*\|_{1,1} - \alpha  \bm{1}^\top\log(\bm{S}_n^*\bm{1}))  + \lambda_1 \bm{Q} + \bm{\Lambda}_2 = \bm{0}, \\
		&\lambda_1(\|\bm{C}_n\bm{S}_n^*-\bm{S}_n^*\bm{C}_n\|_F - \delta_n) = 0, \\
		&\bm{S}_n^* \in \mathcal{S}. 
	\end{aligned}
 \right.
 \end{equation}

Moreover, from the definition of the convex subdifferential we know that $0 \geq \|\bm{C}_n\bm{S}_n^*-\bm{S}_n^*\bm{C}_n\|_F - \langle \bm{Q}, \bm{S}_n^* \rangle$.
	% \begin{align*}
	% 	0 \geq \|\bm{C}_n\bm{S}_n^*-\bm{S}_n^*\bm{C}_n\|_F - \langle \bm{Q}, \bm{S}_n^* \rangle.
	% \end{align*}
    Thus, after taking inner product of both sides of the equation \eqref{equa-optcon-novel} with $\bm{S}_n^*$, it follows that:
	\begin{align*}
		0 &= \sum_{i,j}(\bm{S}_n^*)_{ij} -\alpha m + \lambda_1\langle \bm{Q},\bm{S}_n^* \rangle + \langle \bm{\Lambda}_2,\bm{S}_n^* \rangle \\
        &\geq \sum_{i,j}(\bm{S}_n^*)_{ij} -\alpha m + \lambda_1\|\bm{C}_n\bm{S}_n^*-\bm{S}_n^*\bm{C}_n\|_F + \langle \bm{\Lambda}_2,\bm{S}_n^* \rangle \\
        &= \sum_{i,j}(\bm{S}_n^*)_{ij} -\alpha m + \lambda_1\delta_n,
        % \Rightarrow &\sum_{i,j}(\bm{S}_n^*)_{ij} \leq \alpha m -\lambda\delta_n \leq \alpha m
	\end{align*}
which implies that $\sum_{i,j}(\bm{S}_n^*)_{ij} \leq \alpha m -\lambda_1\delta_n \leq \alpha m$. This completes the proof of the first statement. 
% of the second part of the lemma.
% \end{proof}
% \subsection{Proof of Lemma \ref{lemma-bound-values}}
% \begin{proof}

For the second statement, we first prove that $v_n^*$ and $v^*$ are larger than $\alpha m(1-\log \alpha)$. Define the auxiliary function $g:\mathbb{R}\rightarrow\mathbb{R}$ such that $g(x)\coloneqq x - \alpha\log x$ for any $x\in\mathbb{R}_+$, 
% \[g(x) = x - \alpha\log x\]
whose minimum is attained at $\alpha$. Since for any $\bm{S} \in \mathcal{S}$, 
\[f(\bm{S}) = \sum_{i=1}^mg\left(\sum_{j=1}^m {S}_{ij}\right),\] 
where $f$ is the objective in LogSpecT, it follows that 
\[f(\bm{S}) \geq \sum_{i=1}^mg(\alpha)= \alpha m(1-\log\alpha).\]
This implies that $v_n^*$ and $v^*$ are larger than $\alpha m(1-\log\alpha)$.
Next, we will show
% We next prove that when $\delta_n \geq 2\alpha m\|\bm{C}_n-\bm{C}_\infty\|$, 
$v^*_n \leq v^*$. Consider any optimal solution $\bm{S}^*$ to LogSpecT. We show that it is feasible for rLogSpecT.
\begin{align*}
    \|\bm{C}_n\bm{S}^*-\bm{S}^*\bm{C}_n\| &= \|\bm{C}_n\bm{S}^*-\bm{C}_\infty\bm{S}^*+\bm{S}^*\bm{C}_\infty-\bm{S}^*\bm{C}_n\|\\
    % &\leq 2\|\bm{S}^*\|_F\|\bm{C}_n-\bm{C}_\infty\| \\
    &\leq 2\|\bm{S}^*\|_{1,1}\|\bm{C}_n-\bm{C}_\infty\| 
    \leq 2\alpha m\|\bm{C}_n-\bm{C}_\infty\|
    \leq \delta_n,
\end{align*}
% where the fourth inequality is from \ref{lemma-upbound-solutions} 
where the equality comes from $\bm{C}_\infty\bm{S}^* = \bm{S}^*\bm{C}_\infty$, the first inequality comes from the fact that $\|\bm{X}\bm{Y}\| \leq \|\bm{X}\|_F\|\bm{Y}\|\leq \|\bm{X}\|_{1,1}\|\bm{Y}\|$,
% , the third comes from $\|\bm{X}\|_F \leq \|\bm{X}\|_{1,1}$ 
 the second one comes from the first statement and the last one is due to $\delta_n \geq 2\alpha m\|\bm{C}_n-\bm{C}_\infty\|$.
Hence, $\bm{S}^*$ is feasible for rLogSpecT, which indicates that $v_n^* \leq v^*$. The proof is completed.

\section{
% Truncated Hausdorff Distance: 
Proof of Theorem \ref{thm-main} \& Corollary \ref{coro-conv-sol} }\label{appendix-main}
% by Truncated Hausdorff Distance
\subsection{Truncated Hausdorff distance}
\label{appendix-Truncated Hausdorff Distance}
In this section, we introduce an advanced technique in optimization that is efficient in analyzing the recovery guarantee of robust formulations. Before that, we introduce the concept of truncated Hausdorff distance between two sets. 

\begin{definition}[{Truncated Hausdorff Distance \cite[6.J]{royset2021optimization}}]
    For any $\rho \geq 0$, the truncated Hausdorff distance between two sets $\mathcal{C}$ and $\mathcal{D}$ is defined as:
    \begin{equation*}
        \disth_\rho(\mathcal{C},\mathcal{D}) = \max\{\dist(\mathcal{C}\cap\mathbb{B}(\bm{0},\rho),\mathcal{D}), \dist(\mathcal{D}\cap\mathbb{B}(\bm{0},\rho),\mathcal{C})\}.
    \end{equation*}
\end{definition}

It turns out that the distance between the optimum of two minimization problems can be bounded with the truncated Hausdorff distance of the epigraphs under some conditions. The result is captured in the following lemma.
\begin{lemma}[{\cite[Theorem 6.56]{royset2021optimization}}]
\label{lemma-minimizing-errors}
   Let $\rho \in [0,\infty)$. Suppose that the extended-real-valued functions $f, g : \mathbb{R}^n \rightarrow \overline{\mathbb{R}}$ satisfy
   \begin{itemize}
       \item $\inf f, \inf g \in [-\rho, \rho]$, \label{assumption-minierror-1}
       \item $\argmin f, \argmin g \subseteq \mathbb{B}(\bm{0}, \rho)$.\label{assumption-minierror-2}
   \end{itemize}
   Then, it follows that
   \begin{equation}
       |\inf f - \inf g| \leq \disth_\rho(\epi f, \epi g).\footnote{For a function $f : \mathbb{R}^n \rightarrow \overline{\mathbb{R}}$, its epigraph is defined as
        $\epi f \coloneqq \{(\bm{x},y) \mid y\geq f(\bm{x})\}$.} 
   \end{equation}
   Suppose further that $\varepsilon > 2\disth_\rho(\epi f, \epi g)$, then one has
   \begin{equation}
       \dist(\bm{x}_g^*, \varepsilon\text{-}\argmin f) \leq \disth_\rho(\epi f, \epi g),
   \end{equation}
   where 
   $\varepsilon$-$\argmin f$ is the $\varepsilon$-suboptimal solution set of $f$ that is defined as $\varepsilon$-$\argmin f \coloneqq \{\bm{x} \in \mathbb{R}^n : f(\bm{x}) \leq  \inf f + \varepsilon\}$, and $\bm{x}_g^*$ is a minimizer of $g$.
\end{lemma}
    From the above lemma, we know that if two optimization problems are close enough (in the sense of truncated Hausdorff distance), then the optimum of them should be close to each other. Hence, in order to apply this result, we need to bound the truncated Hausdorff distance in an explicit way, which is solved by the following Kenmochi condition.
\begin{lemma}[{Kenmochi Condition \cite[Proposition 6.58]{royset2021optimization}}]\label{lemma-kenmochi}
    Let $\rho \in [0, \infty)$. Then, for $f, g : \mathbb{R}^n \rightarrow \overline{\mathbb{R}}$ with nonempty
    epigraphs, one has that
    \begin{align*}
        \disth_\rho(\epi f, \epi g) = \inf\left\{\eta > 0 : 
        \begin{aligned}
         \inf_{\mathbb{B}(\bm{x},\eta)}g \leq \max\{f(\bm{x}),-\rho\} + \eta,\ \forall \bm{x}\in [f \leq \rho] \cap \mathbb{B}(\bm{0},\rho)\\
         \inf_{\mathbb{B}(\bm{x},\eta)}f \leq \max\{g(\bm{x}),-\rho\} + \eta,\ \forall \bm{x}\in [g \leq \rho] \cap \mathbb{B}(\bm{0},\rho)
        \end{aligned}
        \right\},
    \end{align*}
    where $[f\leq\rho]\coloneqq\{\bm{x}\in\mathbb{R}^n: f(\bm{x})\leq\rho\}$.
\end{lemma}

\subsection{Proof of Theorem \ref{thm-main}}
\label{appen-proof-thm-main}

Before presenting the proof, we first introduce the following lemma.
\begin{lemma}[{Hoffman's Error Bound \cite{hoffman1952approximate}}]\label{lemma-hoffman}
	Consider the set $\mathcal{S} \coloneqq \{\bm{x}\in\mathbb{R}^n : \bm{Ax} \leq \bm{b}\}$. There exists $C > 0$ such that for any $\bm{x}\in\mathbb{R}^n$, one has
	\[\text{\rm dist}(\bm{x},\mathcal{S}) \leq C\cdot\|(\bm{Ax}-\bm{b})_{+}\|_2.\]
\end{lemma}

For the sake of brevity, we denote
\begin{align*}
	\bar{f}_n(\bm{S}) &\coloneqq \|\bm{S}\|_{1,1} - \alpha\bm{1}^\top\log(\bm{S1}) + \iota_{\mathbb{R}_-}(\|\bm{C}_n\bm{S} - \bm{S}\bm{C}_n\|_F - \delta_n) + \iota_{\mathcal{S}}(\bm{S}), \\
	\bar{f}(\bm{S}) &\coloneqq \|\bm{S}\|_{1,1} - \alpha\bm{1}^\top\log(\bm{S1}) + \iota_{ \{0\} }(\|\bm{C}_\infty\bm{S} - \bm{S}\bm{C}_\infty\|_F) + \iota_{\mathcal{S}}(\bm{S}).
\end{align*}
 Hence, the optimization problem LogSpecT (resp. rLogSpecT) is equivalent to $\inf \bar{f}$ (resp. $\inf \bar{f}_n$).

Now, we aim to use Lemma \ref{lemma-kenmochi} to bound $\disth_\rho(\epi \bar{f}, \epi \bar{f}_n)$. Let $\bm{S} \in \mathcal{S}\cap \mathbb{B}(\bm{0},\rho)$ satisfy 
\[
\bar{f}(\bm{S}) \leq \rho\quad \text{and}\quad \bm{S}\bm{C}_\infty = \bm{C}_\infty\bm{S}.
\]
Then, we know that
\begin{equation*}
        \|\bm{S}\bm{C}_n - \bm{C}_n\bm{S}\|_F \leq 2\|\bm{S}\|_F\|\bm{C}_n - \bm{C}_\infty\| \leq 2\rho\|\bm{C}_n - \bm{C}_\infty\| \leq \delta_n,
\end{equation*}
and consequently $\bm{S}$ is in the domain of $\bar{f}_n$. Then, it follows that for any $\eta > 0$, we have
\begin{equation}\label{equa-dist-epigraph-1}
        \inf_{\mathbb{B}(\bm{S},\eta)}\bar{f}_n \leq \bar{f}_n(\bm{S})  = \bar{f}(\bm{S}) \leq \max\{\bar{f}(\bm{S}),-\rho\},\quad \forall \bm{S}\in [\bar{f} \leq \rho] \cap \mathbb{B}(\bm{0},\rho).
\end{equation}
Before verifying the reverse side of the Kenmochi condition, we first consider the non-emptiness of $[\bar{f}_n \leq \rho]\cap\mathbb{B}(\bm{0},\rho)$. Since 
\[
\delta_n \geq 2\rho\|\bm{C}_n - \bm{C}_\infty\| \geq 2\alpha m\|\bm{C}_n - \bm{C}_\infty\|,
\] 
it follows from Proposition \ref{lemma-bound-solutions} that $\|\bm{S}_n^*\|_{1,1} \leq \alpha m\leq \rho$ and $f^*_n \leq f^*\leq\rho$, 
which implies that $[\bar{f}_n \leq \rho]\cap\mathbb{B}(\bm{0},\rho)$ is nonempty. Let $\bm{S}_n \in [\bar{f}_n \leq \rho]\cap\mathbb{B}(\bm{0},\rho)$. Then, one has that
\begin{equation*}
    \bm{S}_n \in \mathcal{S}\ \ \text{ and }\ \ \|\bm{C}_n\bm{S}_n-\bm{S}_n\bm{C}_n\|_F \leq \delta_n.
\end{equation*}
    Hence, it follows that
    \begin{equation*}
        \|\bm{C}_\infty\bm{S}_n-\bm{S}_n\bm{C}_\infty\| \leq 2\|\bm{S}_n\|_F\|\bm{C}_\infty - \bm{C}_n\| + \| \bm{C}_n\bm{S}_n-\bm{S}_n\bm{C}_n \|_F \leq 2\rho\|\bm{C}_\infty - \bm{C}_n\| + \delta_n. 
    \end{equation*}
    Also, note that there exists $\beta > 0$ such that $(\bm{S}_n\bm{1})_i \geq \beta$ for all $i\in [m]$ as $\bar{f}_n \leq \rho$ and $\|\bm{S}_n\|_{1,1} - \alpha\bm{1}^\top\log(\bm{S}_n\bm{1}) \rightarrow \infty$ when  $\bm{S}_n \rightarrow \bm{0}$.
        Thus, applying Lemma \ref{lemma-hoffman} to the linear system 
    \[
    \tilde{\mathcal{S}} \coloneqq \{\bm{S} \in \mathbb{R}^{m\times m} : \bm{S}\bm{C}_\infty = \bm{C}_\infty\bm{S},\ \bm{S}\in \mathcal{S},\ (\bm{S}\bm{1})_i \geq \beta,\  \forall i \in [m]\}
    \]
    yields that there exists $\tilde{c} > 0$ such that
    \begin{equation*}
        \dist(\bm{S}_n, \tilde{\mathcal{S}}) \leq \tilde{c}\cdot(2\rho\|\bm{C}_\infty - \bm{C}_n\| + \delta_n).
    \end{equation*}
    Hence, there exists $\tilde{\bm{S}}$ in the domain of $\bar{f}$ such that
    \begin{align*}
       \|\bm{S}_n - \tilde{\bm{S}}\|_F &\leq \tilde{c}\cdot(2\rho\|\bm{C}_\infty - \bm{C}_n\| + \delta_n)\ \ \text{ and }\ \ (\tilde{\bm{S}}\bm{1})_i \geq \beta, \ \ \forall i \in [m].
    \end{align*}
    Since the function $\bm{S}\mapsto\|\bm{S}\|_{1,1} - \alpha\bm{1}^\top\log(\bm{S}\bm{1})$ is locally Lipschitz continuous when $(\bm{S}\bm{1})_i \geq \beta$, there exists $L > 0$ such that
    \begin{align*}
        \bar{f}(\tilde{\bm{S}}) = \|\tilde{\bm{S}}\|_{1,1} - \alpha\bm{1}^\top\log(\tilde{\bm{S}}\bm{1}) 
        &\leq \|\bm{S}_n\|_{1,1} - \alpha\bm{1}^\top\log(\bm{S}_n\bm{1}) + L \|\bm{S}_n - \tilde{\bm{S}}\|_F \\
        &= \bar{f}_n(\bm{S}_n) + L \|\bm{S}_n - \tilde{\bm{S}}\|_F \\
        &\leq \bar{f}_n(\bm{S}_n) + L\tilde{c}\cdot(2\rho\|\bm{C}_\infty - \bm{C}_n\| + \delta_n).
    \end{align*}
    Setting $c_1 \ge \max\{1,L\}\cdot\tilde{c}$, one can obtain that for any $\bm{S}_n\in [\bar{f}_n \leq \rho] \cap \mathbb{B}(\bm{0},\rho)$
    \begin{align}
        \inf_{\mathbb{B}(\bm{S}_n,\eta)}\bar{f} &\leq \bar{f}(\tilde{\bm{S}})  
        % \leq \bar{f}_n(\bm{S}_n) + L\tilde{c}\cdot(2\rho\|\bm{C}_\infty - \bm{C}_n\| + \delta_n) \nonumber\\
        \leq \bar{f}_n(\bm{S}_n) + c_1\cdot(2\rho\|\bm{C}_\infty - \bm{C}_n\| + \delta_n) 
        \leq \max\{\bar{f}_n(\bm{S}_n),-\rho\} + \eta, 
\label{equa-dist-epigraph-2}
    \end{align}
    where $\eta \coloneqq c_1\cdot(2\rho\|\bm{C}_\infty - \bm{C}_n\| + \delta_n)$.    
    Combining inequality \eqref{equa-dist-epigraph-1} and \eqref{equa-dist-epigraph-2}, we can conclude that
    \begin{equation}\label{equa-dist-epigraph-3}
        \disth_\rho(\epi \bar{f}, \epi \bar{f}_n) \leq c_1\cdot(2\rho\|\bm{C}_\infty - \bm{C}_n\| + \delta_n).
    \end{equation}

   In order to derive the conclusion (i) and (ii), it remains to check the requirements in Lemma \ref{lemma-minimizing-errors}. Since $\rho \geq \alpha m$, the first statement of Proposition \ref{lemma-bound-solutions} shows that the optimal solutions to $\inf \bar{f}$ and $\inf \bar{f}_n$ lie in $\mathbb{B}(\bm{0},\rho)$. Since $\rho \geq f^*$ and $-\rho \leq \alpha m (1-\log \alpha)$, the second statement of the proposition shows that $\inf \bar{f}, \inf \bar{f}_n\in [-\rho,\rho]$. Hence, 
   % we have checked all the needed conditions. 
   applying Lemma \ref{lemma-minimizing-errors} completes the proof of the first two statements.

   To prove conclusion (iii), we first make the following two claims:
   \begin{enumerate}
       \item[(a)] $\mathcal{S}_0^*\bm{1}$ is a singleton, whose element is denoted by $\bm{S}^*\bm{1}$,
       \item[(b)] 
       For any $\bar{\varepsilon}\in[0,\infty)$, 
       there exists a $ \delta(\bar{\varepsilon}) >0 $
       % (dependent on $M$)
       such that for all $0 \leq \varepsilon \leq \bar{\varepsilon}$ and $\bm{S}_\varepsilon \in \mathcal{S}_\varepsilon^*$, one has that
       \begin{equation}\label{claim}
           \|\bm{S}_\varepsilon\bm{1} - \bm{S}^*\bm{1}\|_2 \leq \delta(\bar{\varepsilon})\cdot\sqrt{\varepsilon}.
       \end{equation}
   \end{enumerate}
Granting these and with the help of Theorem \ref{thm-main}, we can derive that for all $\bm{S}_n^* \in \mathcal{S}^{n,*}$
\begin{equation*}
\begin{split}
    \dist(\bm{S}_n^*\bm{1}, \mathcal{S}_0^*\bm{1}) = \|\bm{S}_n^*\bm{1}-\bm{S}^*\bm{1}\|_2
    & \leq \|\bm{S}_n^*\bm{1}-\bm{S}_{2\varepsilon_n}\bm{1}\|_2 + \|\bm{S}_{2\varepsilon_n}\bm{1}-\bm{S}^*\bm{1}\|_2 \nonumber\\
    &\leq \sqrt{m}\dist(\bm{S}_n^*, \mathcal{S}_{2\varepsilon_n}^*) + \|\bm{S}_{2\varepsilon_n}\bm{1}- \bm{S}^*\bm{1}\|_2 \\
    &\leq \tilde{c}_1\varepsilon_n + \tilde{c}_2\sqrt{\varepsilon_n},
\end{split}
\end{equation*}
where $\tilde{c}_1$, $\tilde{c}_2$ are positive constants, and $\bm{S}_{2\varepsilon_n} \in \mathcal{S}_{2\varepsilon_n}^*$ satisfies 
$\|\bm{S}_n^* - \bm{S}_{2\varepsilon_n}\|_F = \dist(\bm{S}_n^*, \mathcal{S}_{2\varepsilon_n}^*)$ (whose existence is guaranteed since $\mathcal{S}_\varepsilon^*$ is convex and compact).
Hence,
\begin{equation*}
    \dist(\mathcal{S}^{n,*}\bm{1}, \mathcal{S}_0^*\bm{1}) \leq \tilde{c}_1\varepsilon_n + \tilde{c}_2\sqrt{\varepsilon_n}.
\end{equation*}
To proceed, it remains to prove the claims. Define an auxiliary function $h:\mathbb{R}^m\rightarrow \mathbb{R}$ as  $h(\bm{x}) = \sum_{i=1}^mx_i - \alpha \sum_{i=1}^m\log x_i$ for each $\bm{x}\in\mathbb{R}^m_+$. Consider the following optimization problem:
\begin{equation}\label{problem-s1}
    \begin{split}
        &\min_{\bm{x}}\  h(\bm{x}) \\
		&\ \text{s.t.}\ \   \bm{x} \in\{   \bm{S1}\in\mathbb{R}^m\mid  \bm{S} \text{ that is feasible for LogSpecT} \}.
    \end{split}
\end{equation}

For the sake of brevity, denote the $\varepsilon$-suboptimal solution set of \eqref{problem-s1} as $\mathcal{H}_\varepsilon^*$. 
In the remaining part, we will first show that
% It turns out that 
$\mathcal{S}^*_\varepsilon\bm{1}=\mathcal{H}_\varepsilon^*$ and then, by the strict convexity of $h$, the desired two claims hold. 

The first step is to show that the optimal function value 
 of the problem \eqref{problem-s1} satisfies $h^*=f^*$. Since it is obvious that $\Tilde{\bm{x}} = \bm{S}^*\bm{1}$ is feasible for \eqref{problem-s1}, $h^* \leq h(\Tilde{\bm{x}}) = f(\bm{S}^*) = f^*$. Suppose to the contrary that $h^* < f^*$, from the fact that the objective function is coercive and continuous and the feasible set is closed, there exists $\Tilde{\bm{S}}$ such that it is feasible for LogSpecT and $\bm{x}^* = \Tilde{\bm{S}}\bm{1}$, where $\bm{x}^*$ is an optimal solution to \eqref{problem-s1}. Since $h^* = h(\bm{x}^*) = h(\Tilde{\bm{S}}\bm{1}) = f(\Tilde{\bm{S}})$, this contradicts the fact that $f(\Tilde{\bm{S}})\geq f^*$. 
% If $h^* > f^*$, it is obvious that $\Tilde{\bm{x}} = \bm{S}^*\bm{1}$ is feasible to \eqref{problem-s1}. However, $h^* \leq h(\Tilde{\bm{x}}) = f(\bm{S}^*) = f^*$. This is a contradiction. 
Hence, $h^* = f^*$.
Next, we will show that $\mathcal{S}^*_\varepsilon\bm{1} = \mathcal{H}_\varepsilon^*$. Consider any $\varepsilon$-suboptimal solution $\bm{S} \in \mathcal{S}^*_\varepsilon$, i.e., \[h(\bm{S}\bm{1}) = f(\bm{S}) \leq f^* + \varepsilon = h^* + \varepsilon.\]
Hence, $\bm{S}\bm{1} \in \mathcal{H}_\varepsilon^*$ and it implies that $\mathcal{S}^*_\varepsilon\bm{1} \subseteq \mathcal{H}_\varepsilon^*$. On the other hand, for any $\varepsilon$-suboptimal solution $\bm{x} \in \mathcal{H}^*_\varepsilon$, there exists $\bm{S}$ that is feasible for LogSpecT such that $\bm{x} = \bm{S}\bm{1}$. Thus, \[f(\bm{S}) = h(\bm{x}) \leq h^* + \varepsilon = f^* + \varepsilon.\] 
This implies that $\bm{S} \in \mathcal{S}^*_\varepsilon$ and consequently $\mathcal{H}^*_\varepsilon \subseteq \mathcal{S}^*_\varepsilon\bm{1}$. Hence, $\mathcal{H}^*_\varepsilon = \mathcal{S}^*_\varepsilon\bm{1}$.

Since $h$ is strictly convex, its optimal solution set $\mathcal{H}^*_0$ is a singleton. 
% Also, $\mathcal{H}_0^* = \mathcal{S}_0^*\bm{1}$. 
Then, $\mathcal{S}_0^*\bm{1}=\mathcal{H}_0^*$ is a singleton, which proves the first claim.
For the second claim, 
% let $\bm{x}_\varepsilon \in \mathcal{H}_\varepsilon^*$, then
we know that for any $\bm{S}_\varepsilon \in \mathcal{S}_\varepsilon^*$ there exists $\bm{x}_\varepsilon \in \mathcal{H}_\varepsilon^*$ such that
\begin{equation}\label{coro-proof-equation-s2h}
    \|\bm{S}_\varepsilon\bm{1}- \bm{S}^*\bm{1}\|_2 = \|\bm{x}_\varepsilon-\bm{x}^*\|_2,
\end{equation}
where $\bm{x}^* \in \mathcal{H}_0^*$.
The coerciveness of $h$ asserts that $\bm{x}_\varepsilon$ and $\bm{x}^*$ are bounded. This together with the fact that
 $h$ is strongly convex on any bounded set, illustrates that there exists $\mu >0 $ such that
\begin{align}
    &h(\bm{x}_\varepsilon) \geq h(\bm{x}^*) + \langle \nabla h(\bm{x}^*), \bm{x}_\varepsilon - \bm{x}^* \rangle + \frac{1}{\mu}\|\bm{x}_\varepsilon - \bm{x}^*\|_2^2 \geq h(\bm{x}^*) + \frac{1}{\mu}\|\bm{x}_\varepsilon - \bm{x}^*\|_2^2, 
    % &\Rightarrow \|\bm{x}_\varepsilon - \bm{x}^*\|_2^2 \leq \mu(h(\bm{x}_\varepsilon)-h(\bm{x}^*)) \leq \mu\varepsilon,
    \label{coro-proof-coersive}
\end{align}
where the second inequality comes from the global optimality of $\bm{x}^*$.
% and the first-order optimality condition
Combining \eqref{coro-proof-equation-s2h} and \eqref{coro-proof-coersive} gives that
\begin{equation*}
     \|\bm{S}_\varepsilon\bm{1}- \bm{S}^*\bm{1}\|_2 = \|\bm{x}_\varepsilon - \bm{x}^*\|_2 \leq \sqrt{\mu(h(\bm{x}_\varepsilon)-h(\bm{x}^*))}\leq \sqrt{\mu\varepsilon}.
\end{equation*}
This completes the proof of the claims.

\subsection{Proof of Corollary \ref{coro-conv-sol}}
% \begin{proof}
% The corollary is proved by contradiction. If not, 
Suppose to the contrary that there exists a sequence $\{\bm{S}_n^*\}_n$, where the $n$th element is an optimal solution to rLogSpecT with sample size $n$, such that 
\[\dist(\bm{S}_n^*, \mathcal{S}^*_0) \not \rightarrow0.\]
From Proposition \ref{lemma-bound-solutions}, we know that $\{\bm{S}_n^*\}_n$ is bounded, and consequently, has a convergent subsequence. Without loss of generality, we may assume that the sequence itself is convergent and the limiting point is $\bm{S}^*$. Note that 
\begin{align*}
    &\ \ \ \ \ \ \|\bm{C}_n\bm{S}_n^*-\bm{S}_n^*\bm{C}_n \|_F \leq \delta_n, \ \ \bm{C}_n \rightarrow \bm{C}_\infty\ \text{ and } \ \delta_n \rightarrow 0. 
    % &\Rightarrow \  \bm{C}_\infty\bm{S}^* = \bm{S}^*\bm{C}_\infty.
\end{align*}
Hence, $\bm{C}_\infty\bm{S}^* = \bm{S}^*\bm{C}_\infty$.
This indicates that $\bm{S}^*$ is feasible for LogSpecT. 
Then, from Theorem \ref{thm-main}, we know that $f(\bm{S}_n^*) = f^*_n \rightarrow f^*$, which leads to $f(\bm{S}^*) = f^*$ since $f(\cdot) = \|\cdot\|_{1,1} - \alpha\bm{1}^\top\log(\cdot\bm{1})$ is continuous. 
% \begin{equation*}
%     f(\bm{S}_n^*) \rightarrow f(\bm{S}^*).
% \end{equation*}
Together with the fact that $\bm{S}^*$ is feasible, we conclude that $\bm{S}^*$ is an optimal solution to LogSpecT. This further implies that $\dist(\bm{S}_n^*, \mathcal{S}^*_0) \rightarrow 0$, which is a contradiction.

\subsection{Proof of Lemma \ref{lemma-subgaussian}}
% \begin{remark}
    Recall the generative model \eqref{equa-generative-filter}. Since $\bm{w}$ follows a sub-Gaussian distribution, it can be shown that for every $t > 0$, \[\mathbb{P}(\|\bm{x}\|_2 > t)
    % = \mathbb{P}(\|\mathcal{H}(\bm{S})\bm{w}\|_2 > t) 
    \leq \mathbb{P}\left(\|\bm{w}\|_2 > \frac{t}{\|\mathcal{H}(\bm{S})\|}\right) \leq Ce^{-v't^2},\] for some positive constant $v'$, which means that $\bm{x}$ also follows a sub-Gaussian distribution.
% \end{remark}
Thus, due to the sub-Gaussian property, $\|\bm{C}_n-\bm{C}_\infty\|$ can be explictly bounded by 
% \cite{vershynin2012close}. We present the result in 
the following lemma.
\begin{lemma}[{\cite[Proposition 2.1]{vershynin2012close}}]\label{concentration-sub}
	Consider sub-Gaussian, identical, independent random vectors $\bm{x}_1, \bm{x}_2, \ldots, \bm{x}_n \in \mathbb{R}^m$ with $n > m$. Then for all $\varepsilon > 0$, it follows that
	\begin{equation*}
		\mathbb{P}\left(\left\|\frac{1}{n}\sum_{i=1}^{n}\bm{x}_i\bm{x}_i^\top-\mathbb{E}[\bm{x}\bm{x}^\top]\right\|_2 \leq \varepsilon\right) \geq 1- 2e^{2m-l\varepsilon^2n},
	\end{equation*}
for some constant $l>0$.
\end{lemma}
% The above lemma says that if we set 
Setting $\varepsilon^2 = (4/l)\log(2n)m/n$, Lemma \ref{concentration-sub} indicates that with high probability (lower bounded by $1-n^{-1}$), 
\begin{equation*}
	\left\|\bm{C}_n - \bm{C}_\infty\right\| \leq \mathcal{O}\left(\sqrt{\frac{\log n}{n}}\right).
\end{equation*}

\section{Derivations of L-ADMM and Convergence Analysis}
This section includes the details of L-ADMM for rLogSpecT.
\subsection{Proof of Proposition \ref{prop-algo-sub}}
% \begin{proof}
    Note that the minimization problem \eqref{algo-update-zq} is separable for $\bm{Z}$ and $\bm{q}$, and can be split into two subproblems:
    \begin{align}
        &\min_{\bm{Z}\in \mathbb{B}(\bm{0},\delta_n)} \|\bm{C}_n\bm{S}^{(k)} - \bm{S}^{(k)}\bm{C}_n+\bm{\Lambda}^{(k)}/\rho-\bm{Z}\|_F^2,\label{algo-sub-z} \\
        &\ \ \, \, \, \, \min_{\bm{q}}\ -\alpha\bm{1}^\top\log \bm{q} + \bm{\lambda}_2^{(k)\top}(\bm{q}-\bm{S}^{(k)}\bm{1}) + \frac{\rho}{2}\|\bm{q} - \bm{S}^{(k)}\bm{1}\|_2^2.\label{algo-sub-q}
    \end{align}
    For problem \eqref{algo-sub-z}, the optimal solution is the projection of $\bm{C}_n\bm{S}^{(k)} - \bm{S}^{(k)}\bm{C}_n+\bm{\Lambda}^{(k)}/\rho$ onto $\mathbb{B}(\bm{0},\delta_n)$, which is given by
    \begin{align*}
    &\bm{Z}^{(k+1)} = \min\left\{1, \frac{\delta_n}{\|\Tilde{\bm{Z}}\|_F}\right\}\tilde{\bm{Z}} \ \ \text{with} \ \     \tilde{\bm{Z}} = \bm{C}_n\bm{S}^{(k)}-\bm{S}^{(k)}\bm{C}_n+\bm{\Lambda}^{(k)}/\rho. \nonumber 
    \end{align*}
    For problem \eqref{algo-sub-q}, the first-order optimality condition gives
    \begin{align*}
        % &\nabla_{\bm{q}}-\alpha\bm{1}^\top\log \bm{q} + \bm{\lambda}_2^{(k)\top}(\bm{q}-\bm{S}^{(k)}\bm{1}) + \frac{\rho}{2}\|\bm{q} - \bm{S}^{(k)}\bm{1}\|_2^2 = 0 \\
        % \Leftrightarrow 
        &-\alpha 1/\bm{q} + \bm{\lambda}_2^{(k)} + \rho(\bm{q} - \bm{S}^{(k)}\bm{1}) = 0.      
        % \quad \Longleftrightarrow\quad \tilde{\bm{q}} = \frac{1}{\rho}\left(\rho\bm{S}^{(k)}\bm{1}-\bm{\lambda}_2^{(k)}\right),\\
    \end{align*}

This together with the fact that the objective function is convex implies that
\[
\bm{q}^{(k+1)} = \frac{\tilde{\bm{q}}+\sqrt{\tilde{\bm{q}}^2+4\alpha/\rho\bm{1}}}{2}\ \ \text{with}\ \ \tilde{\bm{q}} = \frac{1}{\rho}(\rho\bm{S}^{(k)}\bm{1}-\bm{\lambda}_2^{(k)}).
\]

\subsection{Calculation of $\Pi_{\mathcal{S}} (\cdot)$}
The projection of $\bm{X}$ to $\mathcal{S}$ can be calculated via an optimization problem:
\begin{equation*}
	\begin{split}
		&\min_{\bm{S}}\ \|\bm{X}-\bm{S}\|_F^2 \\
		&\ \text{s.t.}\  \ \, \bm{S}^\top = \bm{S}, \\
		&\ \ \ \ \   \ \ \, S_{ii} = 0,\ i = 1, 2, \ldots,m,\\
		&\ \ \ \ \ \ \ \, S_{ij} \geq 0,\ \forall i, j, 
	\end{split}	 
 \end{equation*}
 which is equivalent to 
 \begin{equation*}
	\begin{split}
		&\min\ \sum_{i<j}\left((X_{ij} - S_{ij})^2 + (X_{ji} - S_{ij})^2\right)\\
		&\ \text{s.t.}\ \ \, S_{ij} \geq 0, \ \forall i <j, \\
         &\ \ \ \ \ \ \ \, S_{ii} = 0, \ \forall i.
	\end{split}
\end{equation*}
Hence
\begin{equation*}	
		(\Pi_{\mathcal{S}}(\bm{X}))_{ij} = 
  \left\{\begin{aligned}
  			&\frac{1}{2}\max\{0,X_{ij}+X_{ji}\}, \quad &&i \neq j,\\
			&0, \quad &&i=j.
		\end{aligned}
	\right.
\end{equation*}

\subsection{Stopping criterion and updating rule of $\rho$}
We follow the procedures in \cite{boyd2011distributed} to update $\rho$ in each iteration. Similarly, we define the primal residual and dual residual as follows:
\begin{align*}
    p_{\rm res}^{(k+1)} &= \sqrt{\|\bm{Z}^{(k+1)} - \bm{C}_n\bm{S}^{(k+1)} + \bm{S}^{(k+1)}\bm{C}_n\|_F^2 + \|\bm{q}^{(k+1)} - \bm{S}^{(k+1)}\bm{1}\|_2^2}, \\
    d_{\rm res}^{(k+1)} &= \rho^{(k)}\left(\bm{C}_n(\bm{S}^{(k+1)}-\bm{S}^{(k)})-(\bm{S}^{(k+1)}-\bm{S}^{(k)})\bm{C}_n + \bm{1}^\top(\bm{S}^{(k+1)}-\bm{S}^{(k)})\bm{1}\right).
\end{align*}
The aim of updating $\rho$ is to control the decaying speed of $p_{\rm res}$ and $d_{\rm res}$ such that their difference is not too large. To this end, we update $\rho$ adaptively following the scheme:
\begin{align*}
    \rho^{(k+1)} \coloneqq \left\{ \begin{aligned}
        &2\rho^{(k)},  &&\text{if} \ p_{\rm res}^{(k+1)} > 5 d_{\rm res}^{(k+1)}, \\
        &\rho^{(k)}/2,  &&\text{if} \ d_{\rm res}^{(k+1)} > 5 p_{\rm res}^{(k+1)}, \\
        &\rho^{(k)},  &&\text{otherwise}.
    \end{aligned}\right.
\end{align*}
When $p_{\rm res}$ and $d_{\rm res}$ are both smaller than the threshold $\varepsilon = 10^{-5}$, we stop the algorithm.
\subsection{Convergence analysis}
\label{sec-convergence-ana}
Define $\bm{D} \coloneqq \Diag(\bm{1}_m^\top,\ldots,\bm{1}_m^\top)\in \mathbb{R}^{m\times m^2}$. Then, $\bm{D}$ satisfies $\bm{D}\text{vec}(\bm{S}) = \bm{S}\bm{1}$ and $\|\bm{D}^\top\bm{D}\| = m$. Denote 
\[
\bm{Q} \coloneqq \tau\bm{I} - \bm{D}^\top\bm{D} - \bm{A}_n^\top\bm{A}_n.
\]
Then the linearized ADMM update \eqref{algo-update-S} of $\bm{S}$ can be written as:
\begin{equation*}
    \min_{\bm{S}}\ L(\bm{S}) + \frac{\rho}{2}\|\text{vec}(\bm{S}) - \text{vec}(\bm{S}^{(k)})\|_{\bm{Q}},
\end{equation*}
where $\|\bm{x}\|_{\bm{Q}} \coloneqq \bm{x}^\top\bm{Q}\bm{x}$.
Since $\tau > m + \|\bm{A}_n\|^2$, we know that $\bm{Q}$ is positively definite. Consequently, by treating $(\bm{Z}, \bm{q})$ as one variable, we can apply Theorem 4.2 in \cite{zhang2011unified} and directly obtain the result.

\section{More Experiments and Discussions on Synthetic Data}
\label{appen-more-synthe}
\begin{figure}[h]
\begin{minipage}[t]{0.49\linewidth}
    \centering		\includegraphics[trim=5 1 50 20,clip = true,width=\textwidth]{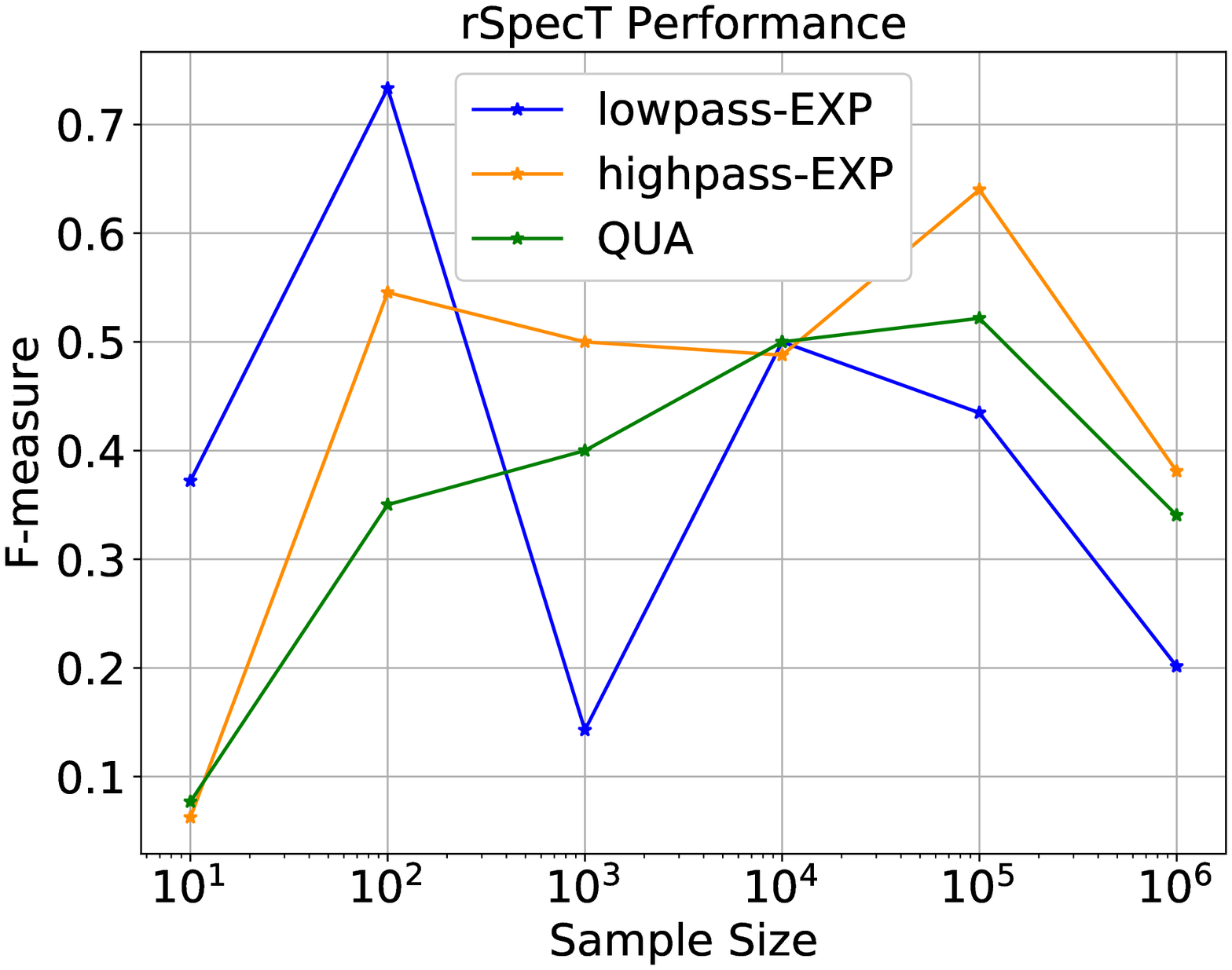}	\caption{\label{fig:syn-rSpecT}Performance of rSpecT on BA graphs.}
\end{minipage}
\begin{minipage}[t]{0.49\linewidth}
    \centering		\includegraphics[trim=10 1 56 30,clip = true,width=\textwidth]{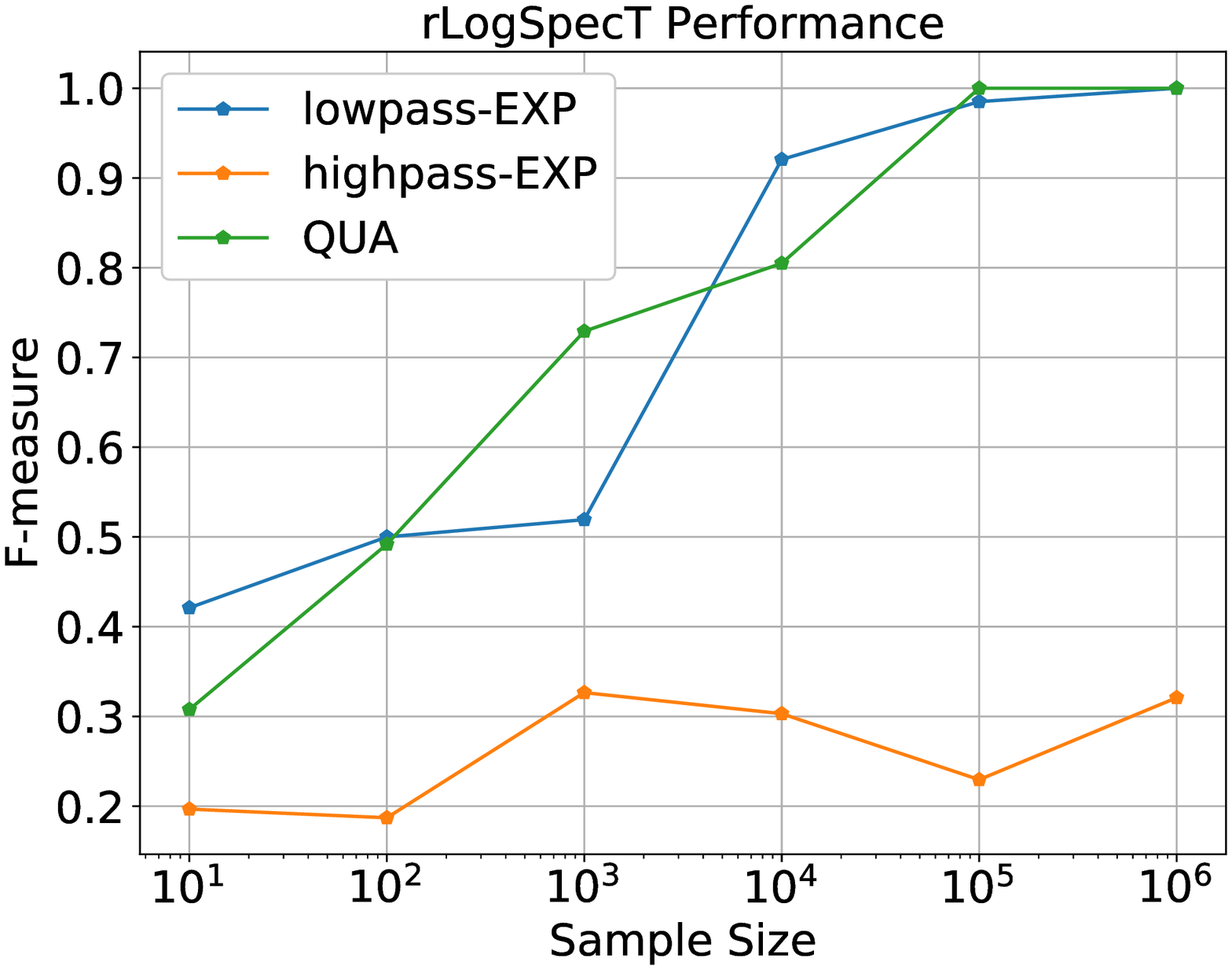}	
    \caption{\label{fig:syn-rLogSpecT}rLogSpecT on ER graphs with $\delta_n = 20\sqrt{\log n/n}$.}
\end{minipage}
\end{figure}

To make a fair comparison between rSpecT and rLogSpecT, we test rSpecT on BA graphs with the same graph filters and the results are reported in Figure \ref{fig:syn-rSpecT}. It is obvious that rSpecT fails in these cases and cannot benefit from the increase in sample size. This is reasonable since SpecT fails on BA graphs as indicated in Figure \ref{fig:syn-idea}, let alone the approximation formulation rSpecT.

We further test rLogSpecT on ER graphs with different numbers of signals observed. The parameter $\delta_n$ is set as $20\sqrt{\log n/n}$ and the results are reported in Figure \ref{fig:syn-rLogSpecT}. The figure shows that for graph filters that are not high-pass, rLogSpecT can achieve nearly perfect recovery when the sample size is large enough. Also, compared with the performance on BA graphs, rLogSpecT works better on ER graphs. This observation is in accordance with the conclusion from Figure \ref{fig:syn-idea} that LogSpecT performs better on ER graphs than BA ones. We further notice that the difference between the low-pass graph filter and the high-pass one is huge. To check the conjecture that rLogSpecT generally performs better on low-pass graph filters, we choose different graph filters $\exp(t\bm{S})$ with $t$ ranging from $-2$ to $2$ and conduct the experiments on ER graphs. When the graph shifting operator is the adjacency matrix, the positive low-pass parameter $t$ corresponds to low-pass graph filters and the negative $t$ corresponds to the high-pass ones \cite{ramakrishna2020user,he2022detecting}. We omit the case when $t = 0$ since this filter does not contain any graph information (note that $\exp(0\bm{S}) = \bm{I}$). 

\begin{figure*}[h]
	\centering		\includegraphics[trim=120 25 100 30,clip = true,width=.98\linewidth]{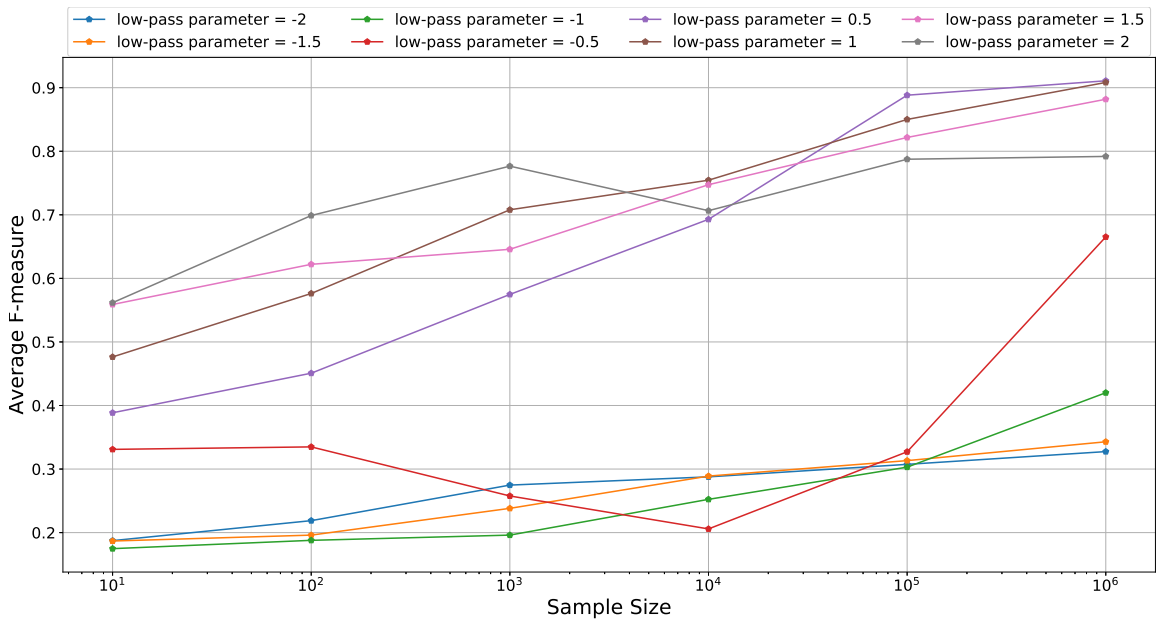}	\caption{\label{fig:syn-effect}Effect of Low-Pass Parameter: different performance of graph filters $\exp(t\bm{S})$ with $t$ ranging from $-2$ to $2$.}
\end{figure*}

We then repeat the experiments for 50 times and report the average results in Figure \ref{fig:syn-effect}. The comparison between the performance of low-pass graph filters and high-pass graph filters indicates that the low-pass graph filters generally outperforms the high-pass ones. A closer look at the results shows that the performance grows faster when the absolute value of $t$ is smaller. And eventually, the graph filter with smaller absolute value of $t$ prevails. This observation is interesting since Figure \ref{fig:syn-idea} indicates that the choice of graph filters has few impacts on the model performance. One explanation is that both low-pass graph filters and high-pass graph filters attenuate some frequencies of the graph and the larger absolute value of $t$ leads to the more loss of information carried by finite signals. 

%%%%%%%%%%%%%%%%%%%%%%%%%%%%%%%%%%%%%%%%%%%%%%%%%%%%%%%%%%%%%%%%%%%%%%%%%%%%%%%
%%%%%%%%%%%%%%%%%%%%%%%%%%%%%%%%%%%%%%%%%%%%%%%%%%%%%%%%%%%%%%%%%%%%%%%%%%%%%%%
\end{appendices}

\end{document}